# A multi-level interpretable sleep stage scoring system by infusing experts' knowledge into a deep network architecture

Authors: Hamid Niknazar, Sara C. Mednick
Department of Cognitive Sciences, University of California, Irvine

**Abstract:** In recent years, deep learning has shown potential and efficiency in a wide area including computer vision, image and signal processing. Yet, translational challenges remain for user applications due to a lack of interpretability of algorithmic decisions and results. This black box problem is particularly problematic for high-risk applications such as medical-related decision-making. The current study goal was to design an interpretable deep learning system for time series classification of electroencephalogram (EEG) for sleep stage scoring as a step toward designing a transparent system. We have developed an interpretable deep neural network that includes a kernel-based layer based on a set of principles used for sleep scoring by human experts in the visual analysis of polysomnographic records. A kernel-based convolutional layer was defined and used as the first layer of the system and made available for user interpretation. The trained system and its results were interpreted in four levels from microstructure of EEG signals, such as trained kernels and effect of each kernel on the detected stages, to macrostructures, such as transition between stages. The proposed system demonstrated greater performance than prior studies and the results of interpretation showed that the system learned information which was consistent with expert knowledge.

## 1. Introduction

Deep learning (DL) systems could propose results comparable to and in some cases surpassing human expert performance [1]. However, one of the main criticisms towards DL networks is related to their 'black-box' nature and their lack of transparency and interpretability for users [2]. Using DL networks as an end-to-end system could improve the performance of automatic systems, but the challenge is to explain the decision-making process based on the characteristics of the input data. A lack of this level of interpretability creates resistance to accept the results, particularly for applications that require a high degree of safety, such as medical diagnostics [3]. One solution is to design a system that provides a better understanding of the working mechanism of DL systems to make it interpretable and understandable to the user [1]. Prior studies have proposed approaches to visualize, explain, and interpret DL models [4]–[6]. However, generally, there is a tradeoff between interpretability and performance [7]. Using expert's knowledge and imitating the process which experts follow for performing the task in designing a DL model may be a useful approach for developing an interpretable DL system, especially in cases that provide a standard model with accepted patterns of input data, such as with polysomnography.

Sleep is a vital process conserved across all species. Sleep stage scoring provides an initial and tangible illustration of how the brain state changes across sleeping time [8], and is used diagnostically to evaluate for sleep disorders as well as to measure the effect of sleep on cognition [9]–[13]. Sleep scoring experts use electrophysiological signals such as electroencephalogram (EEG), electrooculogram (EOG), electromyogram (EMG), and electrocardiogram (ECG) to visually score sleep stages according to the gold standards, Rechtschaffen and Kales (R&K) and AASM modified R&K criteria [14], [15]. Visually scoring sleep stages is complex, time-consuming, and subjective [16]. For example, Danker-Hopfe et al. reported that interrater reliability for sleep scoring according to both R&K and AASM was 80.6% (Cohen's kappa = 0.68) and 82.0% (Cohen's kappa = 0.76), respectively [17]. In another study using 1800 30s epochs, the overall average agreement of 2500 scores with 3 or more years of experience was reported as 82.6% [18]. Therefore, designing a robust and accurate automatic sleep stage scoring system is warranted.

Many sleep stage scoring machine learning algorithms have been proposed over the past four decades [19], including traditional (shallow) and deep learning techniques. Shallow learning based algorithms on a range of features, including temporal [20]–[22], frequency [23]–[25], time-frequency [22], [26], [27] and nonlinear features [22], [28], as well as a range of classifiers including artificial neural networks [29], k-means [30], support vector machines [31] and ensemble classifiers [32]–[34]. On the other hand, DL techniques during the past years have surpassed the heat of machine learning models. Deep learning has shown its potential and effectiveness; especially convolutional neural networks (CNN) in computer vision [35]–[37] and recurrent neural networks (RNN) in natural language processing (NLP) [38]–[40]. Also, DL showed its proper efficiency in electrophysiological signal processing [41]–[46]. In recent years, DL techniques have supported automated sleep stage scoring using CNNs as an end-to-end system [47]–[55] and RNNs to better model the transitions between stages, as well as including many epochs for decision-making [48], [56]–[58].

In the visual sleep stage scoring, experts try to follow criteria, such as the R&K manual, to perform the scoring and there is some information about how the decision can be made and based on what kind of characteristics observed on the PSG records. For example, experts look for different well-known waveform patterns including spindles, k-complex, slow oscillations, alpha and theta waves in EEG signals, eyes movement reflected in EOG signals, and body movements. Also, they consider not just each 30s epochs individually but also make decisions staging decisions by considering the surrounding epochs. Therefore, this study tried to infuse some main experts' knowledge about the scoring criteria in architecture designing of a DL system for automatic sleep stage scoring. Also, the proposed system could be interpretable after training and provide some learned knowledge from the data.

## 2. Related Works
### 2.1. Automatic sleep stage scoring

In recent years, in corresponding to the emerging success of deep learning based application in all aspects, significant progress has been made towards automatic sleep stage scoring. The availability of big sleep datasets [59], [60] has enabled deep learning methods capability of learning high-dimensional features from sleep databases directly to solve the sleep stage scoring problem [49], [56], [61], [62]. The attempts to use deep learning in this application are spread from vanilla architectures such as CNNs [47], [49], [63], [64], and RNNs [65] to more complex architecture and approaches such as [66] which used a CNN based on a modified ResNet-50 to extract sleep-related features and a two-layer bidirectional long short-term memory (Bi-LSTM) to learn the transition rules between sleep stages, and XSleepNet [61] which used a sequence-to-sequence model and jointed representation from both raw signals and time-frequency images. The existing works can be categorized into two categories base on the input of the network. The first processes raw signals as 1-dimentional input [47]–[50], [58] and the second uses 2-dimentional images which are produced mostly by time-frequency transformations such as short-time Fourier transform and wavelet [56], [57], [64], [65]. Although time-frequency transformations, in general, is considered as a higher level representation, but in some cases, using raw signals could lead to better performance. For example, there are some pre-defined waveform of raw EEG signals, such as slow oscillation and spindles, which modeling them in the architecture may improve performance. Gabor is one of the functions which it is shown that is capable of modeling EEG patterns [67]–[69].

### 2.2. Gabor kernel

Based on the sleep stage scoring manuals (R&K and AASM) some EEG frequency sub-bands, existence and density of some specific EEG waveforms (such as spindles, slow oscillation (SO), and K-complex) play important roles in visual scoring process. Gabor function, based on its definition, is capable of modeling signals/images and their patterns based on their frequency, phase and envelope. Therefore, it has been used widely for modeling and transforming signals and images such as extracting feature from electrophysiological signals [70]–[77]. Also, Gabor has been used as a kernel in deep neural network architecture to define a new layer to improve feature extraction process in different classification application such as speech recognition [78] which used Gabor as filter kernels to extract features from power-normalized spectrum and image classification [79]–[82]. Using Gabor kernel as a filter in first layer of a deep neural network architecture may increase the interpretability [83]–[85] as some patterns can be represented using the optimized kernels after the training which are more compatible with the human knowledge.

### 2.3. Deep learning and interpretation

Interpretability is one the biggest challenges that deep learning is faced with. The studies which tried to address this problem (specially in image classification) can be categorized into two groups. In the first category, the methods visualize sensitivity of the outputs to pixels areas on the input image such as saliency map [86], integrated gradients [87], CAM [88], and Grad-CAM [89]. In this category there is a need to pass an input data and the methods highlight areas on the input image as important pixels relevant to the output. The other category tries to represent the network's learnt knowledge after the training process. For example, [90] proposed a method to walk through the convolutional layer end explore what kind of information, and what level of details of the input image were extracted and used by the network to make a decision.

### 2.4. Sleep stage scoring system interpretation

Some studies tried to design interpretable deep learning based system to explain the learnt knowledge and show which part of the inputs had most effect on the decision making. For example, Pathak et al. [91] proposed a system including a CNN fed by three modalities (EEG, EOG and EMG signals) to extract spatio-temporal features followed by a Bi-LSTM to extract sequential information to score sleep stages. They used occlusion method [90] to show which of the modalities, what time interval and what frequency sub-band of EEG signal have the most effect on the predicted sleep stage. Vilamala et al. [92] used transfer learning of a pretrained VGG network to classify time-frequency domain of EEG signal and by calculating the sensitivity of the output to the input pixels found important time and frequency intervals of EEG signal. In [93] a new patching method is proposed to split spectrogram of EEG signal into different time and frequency patches and an attention architecture is used to score sleep stages. Then the attention visualization used to describe which patches had most effect on the predicted sleep stage. By combining deep learning models with expert defined rules, Al-Hussaini et al. [94] was able to make simple interpretable models like a shallow decision tree which can be used to interpret how the system used time and frequency information to predict sleep stages. Recently, Baek et al. [95] proposed an automatic sleep scoring model based on intrinsic oscillations in a single channel EEG signal using bivariate empirical mode decomposition to extract the intrinsic mode components and an attention mechanism to provide weights to the components depending on their significance to sleep scoring as interpretation. Although all these studies have provided interpretation in some levels, but mostly they have used methods which were introduced for image classification. It is noteworthy that in many image classification tasks, there is no exact description of how human make the decision and by using which exact features and properties of an image. But, in the sleep stage scoring task, the experts know the exact properties of each class and also it is standardized in the manuals such as AASM and R&K.

### 3. Problem Definition

Although using the well-known deep learning architecture such as CNN and attention may provide high performance automatic sleep stage systems, but interpreting these systems would not be close to the expert knowledge and compatible to the sleep stage scoring manuals. Therefore, infusing sleep experts' knowledge to the network architecture may have a two main benefits. First, as the decision-making process can be more similar to the real-world process that the experts use for sleep stage scoring, the performance would be improved, and next, the interpretability of the trained system may increase and the interpretation would be closer to the experts knowledge. Therefore, we designed a system to score sleep stages which follows the main principles that the sleep experts use during the sleep stage scoring process by adding a new kernel-based layer as the first layer and proposed multi levels of interpretation to describe the learnt knowledge by the system and the decision-making process.

## 4. Interpretable Model

One primary criterion used by experts is the identification of specific EEG waveforms and frequency ranges, including slow oscillations (~1 Hz), alpha (8-13 Hz), theta (3-7 Hz), spindles (15-18 Hz bursts of activity in a spindle shape), K-complexes (large biphasic waves), and movement artifacts, along with slow and rapid eye movements observable in the EOG. To mimic specific waveforms identification by human expert, instead of using common CNN architecture as a feature extractor, we defined a trainable kernel-based layer using Gabor function (Eq. 1) that can be fitted to the mentioned waveforms by changing the parameters.

$$G(t^*) = e^{-\pi \frac{(t^* - u)^2}{|\sigma|}} \cos(2\pi f t^*)$$

The Gabor function with these three parameters ($u$ as time offset, $\sigma$ as the standard deviation of Gaussian envelope and $f$ as the main frequency) can model EEG and EOG waveforms. Figure 1 Shows $G(t^*)$ in different parameter sets which follows the important sleep related EEG waveforms. Based on the properties of the EEG and EOG waveforms the time margin of the Gabor functions was set to 2 seconds ($-1s \leq t^* \leq 1s$). By using the Gabor function, the Gabor layer (GL) was defined by Eq. 2.

$$GL^i(t) = G^i(t^*) \star X(t)$$

Where $GL^i(t)$ is i-th channel of the output signals of the layer. $G^i(t^*)$ is a Gabor function waveform with a specific set of parameters, X is the input signal (EEG/EOG), and $\star$ is the cross-correlation operation. The $GL$ layer is similar to a one-dimensional convolutional layer with the Gabor function as its filters which the parameters of the Gabor kernels are trainable. After applying the signal, the output of the $GL$ represents the magnitude of the specific waveforms (Gabor kernels) across the time. Using Gabor function in the first layer of the proposed model can increase the interpretability of the model after training in some ways which is described in section 4.2.

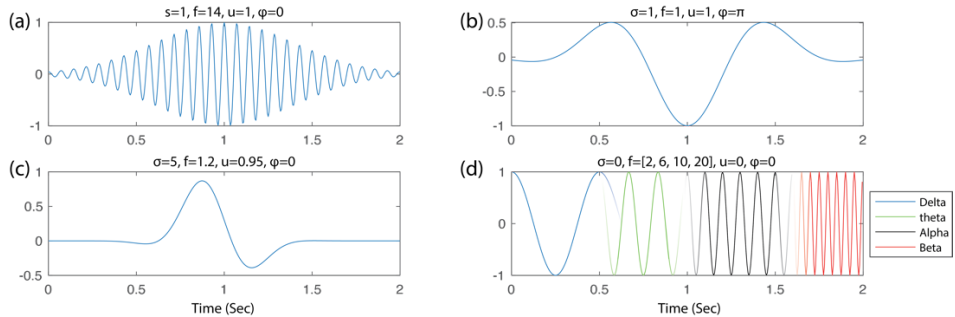

Figure 1. Gabor function waveform in some different parameter sets to produce sleep related EEG patterns. a) Gabor waveform in a parameter set to follow spindle properties. a) Gabor waveform in a parameter set to follow sleep slow oscillation properties.

*a) Gabor waveform in a parameter set to follow K-complex properties. a) Gabor waveform in four different parameter sets to follow main EEG frequency sub-bands.*

In addition, experts score signal epochs by considering the surrounding epochs. Therefore, we designed the network architecture based on two levels of the decision-making process. In the first level, a network produces the probability of each stage based on the waveforms in a single epoch. The second network learns sleep stage scoring based on the probabilities of different stages of each epoch and its surrounding epochs. The details of these two networks are described in the following section.

### 4.1. Network Architecture

Figure 2 shows the architecture of the two networks. The single-epoch network consists of two GL in the first layer, one for EEG signal with 32 kernels and the second for EOG signal with 8 kernels. The output of each GL is N one-dimensional signals where N is the number of kernels. Each output signal represents the cross-correlation of a Gabor kernel and the input signal. The second layer, the mixing layer, is a one-dimensional convolutional layer with the kernel size of 1. The mixing layer mixes the output of GLs to feed the following CNN architecture for feature extraction and classification. The outputs of the single-epoch network for n-th epoch of EEG and EOG data ($O_n$), four previous and next epochs ($O_{n-4}$ to $O_{n-1}$ and $O_{n+1}$ to $O_{n+4}$) are used as the input of the multi-epoch network as the second network (the multi-epoch network). The multi-epoch network consists of two long short-term memory (LSTM) layers and a fully connected layer (FC) for classification. The first LSTM ($LSTM_1$) feeds the output of the first level network in the forward direction and the second LSTM ($LSTM_2$) feeds the output of the first level network in the backward direction (Figure 2).

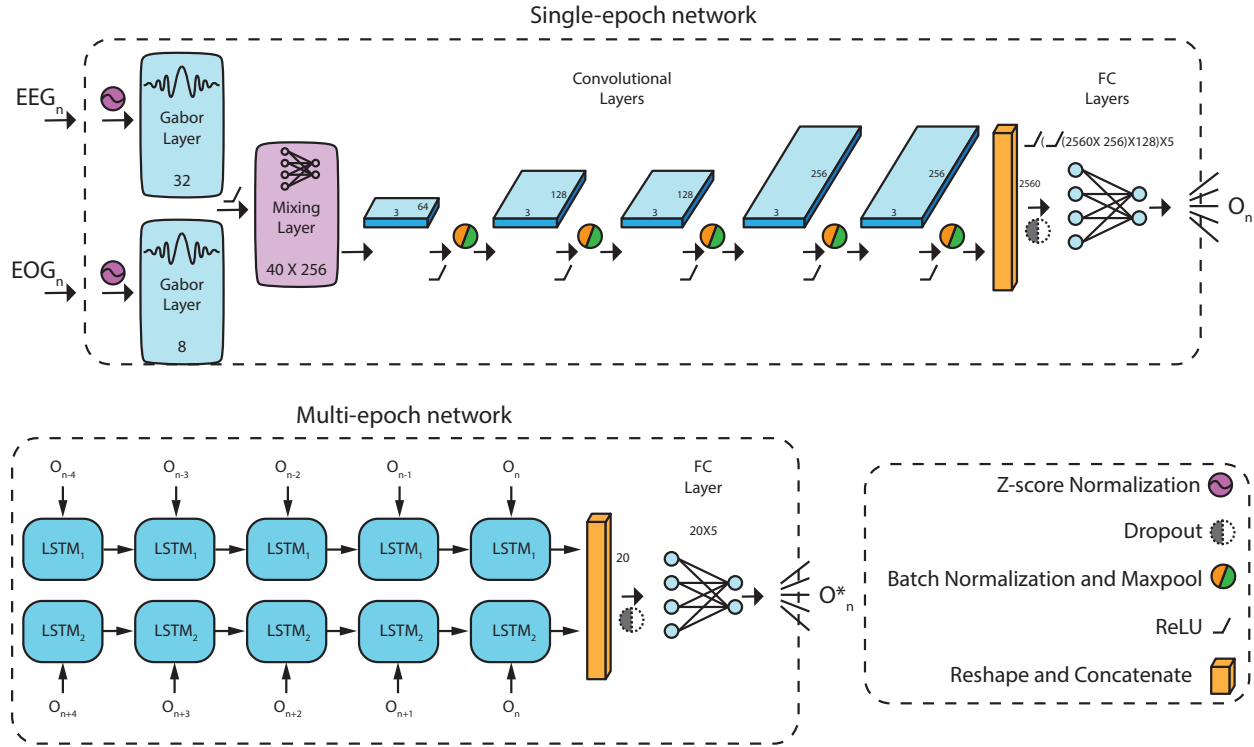

*Figure 2. The architecture of the proposed method for sleep staging including single- and multi-epoch networks. The single-epoch network scores each 30s epoch of EEG and EOG signals by using Gabor kernel waveforms and convolutional layers. The multi-epoch network uses the outputs of single-epoch network for sequential inputs to score sleep stages using two LSTM layer.*

Table 1 shows the details of the single-epoch network and the properties of each layer. The LSTM layers in the multi-epoch network consist of two layers with a hidden state size of 10. The output of the single-epoch network for the last 4 epochs to the next 4 epochs are considered as the inputs of the multi-epoch network which feed the two LSTMs in forward and backward directions (Figure 2).

*Table 1. The architecture details of single-epoch network.*

|    | Layer name | Properties |
|----|------------|------------|
| 1  | Z-score Normalization | |
| 2  | EEG and EOG GL | EEG $N_{Kernels}$=32, EOG $N_{Kernels}$=8 |
| 3  | ReLU | |
| 4  | Mixing layer (1D CL) | Size kernel=1, $N_{filters}$=256, Stride=1 |
| 5  | 1D CL | Size kernel =3, $N_{filters}$=64, Stride=1 |
| 6  | ReLU | |
| 7  | 1D MaxPool | Size kernel =3, Stride=3 |
| 8  | Batch normalization [96] | |
| 9  | 1D CL | Size kernel =3, $N_{filters}$=128, Stride=1 |
| 10 | ReLU | |
| 11 | 1D MaxPool | Size kernel =3, Stride=3 |
| 12 | Batch normalization | |
| 13 | 1D CL | Size kernel =3, $N_{filters}$=128, Stride=1 |
| 14 | ReLU | |
| 15 | 1D MaxPool | Size kernel =3, Stride=3 |
| 16 | Batch normalization | |
| 17 | 1D CL | Size kernel =3, $N_{filters}$=256, Stride=1 |
| 18 | ReLU | |

| 19 | 1D MaxPool | Size kernel =3, Stride=3 |
| 20 | Batch normalization | |
| 21 | 1D CL | Size kernel =3, N$_{filters}$=256, Stride=1 |
| 22 | ReLU | |
| 23 | 1D MaxPool | Size kernel =3, Stride=3 |
| 24 | Batch normalization | |
| 25 | Dropout | P=0.5 |
| 26 | Full connected | N$_{neurons}$=256 |
| 27 | ReLU | |
| 28 | Full connected | N$_{neurons}$=128 |
| 29 | ReLU | |
| 30 | Full connected | N$_{neurons}$=5 |

### 4.1.1. Training Process

The training process consists of two main steps. The single-epoch network was trained using 30-second EEG and EOG epochs. Then, single-epoch network outputs of the training set were used to train the multi-epoch network. Based on the nature of the sleep cycle and different distributions of the sleep stages, a probabilistic approach was used for making mini-batches to avoid biasing the network's output to stages with a greater number of epochs. In the training process, the minibatch size was considered as 16 and the minibatch was constructed based on randomly selected epochs using a uniform probability for stages.

The trainable parameters in the single-epoch network consist of parameters of the Gabor kernels in GLs, convolutional and full connected layers parameters. The cross Entropy was used as the loss function which combines the negative logarithm likelihood and logarithm of SoftMax function (Eq. #).

$$loss(O, class) = -\log\left(\frac{exp(O[class])}{\sum_k exp(O[k])}\right)$$

where $O[k]$ is the k-th element of the output vector of the network and $class$ is the score which is labeled by the experts. The Adam algorithm [97] with an initial learning rate of 0.000625 (0.01/minibatch size) was used in the training to optimize the trainable parameters. The learning rate was decreased after every 5000 iterations and the validation process was applied every 1000 iterations to avoid overfitting.

After training the single-epoch network, the outputs ($O_n$) of all training data were used as training data to train the multi-epoch network. The same loss function, learning rate initialization and adjustment, and optimization algorithm were used for training the multi-epoch network.

### 4.1.2. Evaluation metrics

To evaluate the performance of the proposed method of sleep stage scoring. We used recall, precision, accuracy, F1-score, macro F1-score (MF1), and Cohen's kappa. Among these measures, recall, precision, and F1-score showed the performance of the method with respect to each sleep stage separately. These measures are defined as follow:

$$recall = \frac{TP}{TP + FN}$$

$$precision = \frac{TP}{TP + FP}$$

$$F1 - score = 2\frac{precision \times recall}{precision + recall}$$

Where, for a specific stage S, $TP$ (True Positive) is the number of 30s epochs in stage S which correctly classified in stage S, $FN$ (False Negative) is the number of 30s epochs in stage S which are not classified in stage S, and $FP$ (False Positive) is the number of 30s epochs which are classified in stage S but do not belong to stage S. The rest measures including accuracy, MF1, and Cohen's kappa were considered global performance measures. Accuracy value represents the ratio of correctly labeling (Eq. #), MF1 is average of F1-score across all stages, and Cohen's kappa calculates the degree of agreement in classification over that which would be expected by chance (Eq. #).

$$\text{Accuracy: Acc} = \frac{\sum_{stages} TP}{Number\ of\ all\ samples}$$

$$\text{Kappa: } \kappa = \frac{\bar{P} - \bar{P_e}}{1 - \bar{P_e}}$$

Where $P_e$ is the agreement probability and $\bar{P}$ is the proportion of observed agreements (Accuracy).

5. **Interpretation Approach**

Using the Gabor function with trainable parameters, as the kernels in the first layer, forces the single-epoch network to learn the waveforms that are associated with the sleep scoring process. Therefore, the optimized waveforms after the training process can be used to interpret the learned knowledge of the network.

The GL output consists of applied Gabor kernels on the input signals which can show the correlation between the input signal and the kernel waveform, representing how much a kernel waveform exists in the input signal across the time. Although the output of the GL is meaningful to interpret, the amplitudes in each output time series are not comparable to other outputs. The subsequent layers of the network and the GL output values cannot represent how much each waveform assists to have the network output. Local gradients of the GL's outputs based on the target output can be used as the normalization weights. Therefore, the sensitivity of the network output to GL's outputs time series $(Sen(t)_{GL^i(t) \to O[class]})$ can be measured by computing derivation of the considered output $(O[class])$ based on each sample the Gabor layer outputs $(GL^i(t))$ (Eq. #).

$$Sen(t)_{GL^i(t) \to O[class]} = \frac{dO[class]}{dGL^i(t)}$$

The relation between the GL's outputs time series and the network output is highly complex, nonlinear, and depends on the inputs. We considered the derivation equation from the linear approximation as the sensitivity of the output to each GL output. The sensitivity only shows the dependency of output changes to the GL outputs changes. Using a mixture of the GL output values along with the sensitivity represents the functional effect of the GL outputs on the network output. Therefore, we used the sum squared multiplication of GL outputs by the positive values of sensitivity to quantify the power of the positive functional effect of each Gabor kernel to the network output by the given input (Eq. #).

$$Eff_{GL^i \to O[class]}^{X(t)}(t) = GL^i(t) \times Sen(t)_{GL^i(t) \to O[class]} \times \theta(Sen(t)_{GL^i(t) \to O[class]})$$

$$\overline{Eff}_{GL^i \to O[class]}^{X(t)} = \sum_t (Eff_{GL^i \to O[class]}^{X(t)}(t))^2$$

Where $\theta(.)$ is a step function and $\overline{Eff}_{GL^i \to O[class]}^{X(t)}$ is the positive functional effect of i-th Gabor kernel on the *class*-th element of the output by the given input $X(t)$ (which can be EEG or EOG).

Different levels of interpretation were defined based on the designed architecture, $Eff_{GL^i \to O[class]}^{X(t)}(t)$ time series ($Eff(t)$), and $\overline{Eff}_{GL^i \to O[class]}^{X(t)}$ ($\overline{Eff}$) from the top (overall) to the bottom (detailed) (Figure 3). At the first level, staging process level, based on the considered logic in single-epoch network design and using Gabor kernels in the first layer, after an efficient training process, we assumed that the system would find the optimized parameter sets of the Gabor kernels allowing for interpretation of the impact of each waveform in EEG signals on the detected sleep stages. Therefore, by weighted averaging the $\overline{Eff}$ for all Gabor kernels over all test epochs, the impact of each kernel (*i*-th kernel) on the sleep staging process can be quantified as $Eff_i$ as a qualitative measure (Eq. #).

$$Eff_i = \sum_j \frac{1}{N_j} \sum_{X(t)} \left[ \overline{Eff}_{GL^i \to O[class_X]}^{X(t)} \delta(class_X - j) \right]$$

Where $N_j$ is total number of the test epochs in sleep stage $j$, $\delta(.)$ is the unit impulse function and $O[class_X]$ is the real sleep stage of the relative input signals ($X(t)$). The next level focuses on each sleep stage separately (stages level). To measure the effect of the kernels on the sleep stage scoring process in a quantitative approach, the weighted number of the times that each Gabor kernel had the highest effect on the selected sleep stage were defined as $\#_i$ for the *i*-th Gabor kernel as Eq.#.

$$\#_i = \sum_j \frac{1}{N_j} \sum_{X(t)} \left[ \delta \left( \operatorname*{argmax}_k \left( \overline{Eff}_{GL^k \to O[class_X]}^{X(t)} \right) - i \right) \delta(class_X - j) \right]$$

Where $\#_i$ has linear relation to the number of times that $i$-th kernel has the highest positive functional effect ($\overline{Eff}$) on the sleep staging process. Similarly, the impact of each waveform on each sleep stage can be measured in both quantity and quality. By averaging the values of $\overline{Eff}$ for each kernel in each sleep stage, the overall impact of each kernel waveform on each specific sleep stage can be quantified as $Eff_i^j$ (Eq. #).

$$Eff_i^j = \frac{1}{N_j} \sum_{X(t)} \left[ \overline{Eff}_{GL^i \to O[class_X]}^{X(t)} \delta(class_X - j) \right]$$

After computing $\overline{Eff}$ for each kernel in each epoch and sorting these values for each epoch in each stage, we computed the number of times that each kernel had the highest impact on choosing the selected sleep stage as $\#_i^j$ (Eq.#).

$$\#_i^j = \sum_{X(t)} \left[ \delta \left( \underset{k}{\mathrm{argmax}} \left( \overline{Eff}_{GL^k \to O[class_X]}^{X(t)} \right) - i \right) \delta(class_X - j) \right]$$

Where $\#_i^j$ represent the number of times that $i$-th kernel has the highest positive functional effect ($\overline{Eff}$) in sleep stage of $j$.

The next level of interpretation focuses on the signals in two different approaches (signals level). By comparing the output of the single- and multi-epoch networks, we can observe when the system considers the surrounding epochs for scoring. Also, by averaging the $\overline{Eff}$ values for all kernels of EEG and EOG signals separately it is possible to evaluate the importance of EEG and EOG signals to detect the current sleep stage. At the final level (EEG across time level), $Eff(t)$ of EEG related kernels is used to demonstrate when and how much each waveform impacts the selected sleep stage.

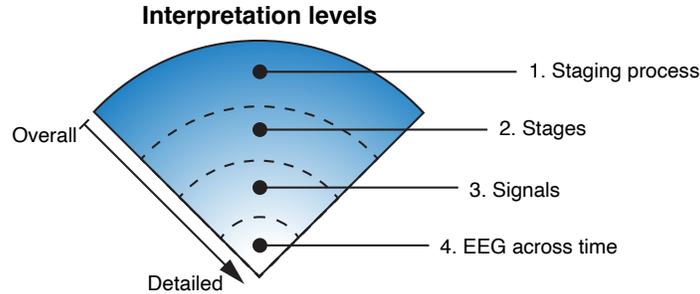

*Figure 3. Overview of levels of interpretation. The first level contains interpretation about the Gabor kernels. The second level contains extracting information about each stage separately. The third level focuses on using single epoch or multi epochs signals and EEG vs. EOG signals. The last level interprets the Gabor waveform impact on detected sleep stage across time.*

## 6. Experiments
### 6.1. Data

Three PSG datasets were used in this study including Physionet Sleep-EDF [98], [99], Physionet EDF-20 and DREAMS [100]. Physionet Sleep-EDF dataset is considered as the main dataset in this study as includes greater number of records and has two nights records for each subject which cause possibility to evaluate the proposed method in different approaches. This dataset was utilized to show the capability and generalizability of the proposed method. Also, only this dataset has been used in the interpretation experiments. The two other datasets were used to improve comparability of the proposed method's performance in comparison to other studies as these two datasets have been used widely in other studies.

*EDF Expanded dataset_* The Cassette records of Physionet Sleep EDF Expanded dataset were utilized including 153 PSG recordings from 78 healthy subjects aged 25-101, without any sleep-impacting medication. Each record consists of two EEG channels (Fpz-Cz and Pz-Oz), one horizontal EOG channel which were each sampled at 100 Hz, and one EMG channel which was sampled at 1Hz. This study used the Fpz-Cz EEG channel and the EOG channel of the records. Except for three subjects, there were about 20 hours each recorded during two subsequent day-night periods. The recordings were manually scored by well-trained technicians according to the R&K manual into six different stages including wake, S1, S2, S3, S4, and REM. However, in this paper, we mixed two sleep stages S3 and S4 into one stage as SWS as is consistent with current scoring practice.

Three different approaches were considered to divide the dataset into training, validation, and test sets. First, the night-holdout k-fold cross validation approach, recording of 75 subjects with two subsequent recordings, were split into 120 recordings for training and 30 recordings for test k=5 times with the condition of having only one of the subsequent recordings in the test set and covering all the subjects in the testing process. The rest of the three recordings were considered the validation set. This approach was designed to evaluate the proposed sleep staging system for new records from subjects in which the system is trained by data of same subjects. Second, the subject-holdout k-fold cross validation approach, two recordings from 60 subjects (about 80% of the subjects) were used for training, recordings from rest of the subjects which had two recordings (15 subject, about 20% of subjects) were used as the test set. Data from three other subjects that did not contain subsequent day-nights were considered as the validation set. In this approach, the system was evaluated by recordings from subjects which the system did not use their data in the training process. In the third record-holdout approach, 15 randomly selected recordings (about 10% of the dataset) were considered for the test, 3 recordings for validation, and 135 recordings for training (about 90% of the dataset). This approach was used only to explain the knowledge which the system was learned in the training process to show the interpretability of the system. Table 2 shows the average number and variation of epochs in each of the approaches for each sleep stage.

*EDF-20 dataset _* The EDF-20 dataset includes 20 PSG records from the Cassette records of Physionet Sleep EDF Expanded dataset with the same properties of sampling rate and included signals. This dataset was used to compare performance of the proposed method in sleep stage scoring with other studies. The leave-one-out (LOO) method is the most used validation method on this dataset in which one recording is used as the test data ad the rest records are used for training process and this process repeats for all of the records to have all the records in the test set and the general performance is average of all of the tests. Table 2 shows the average number and variation of epochs in each of the sleep stages for the train and test sets.

*DREAMS dataset* _ The DREAMS database includes eight different datasets but in this study the DREAMS subjects dataset was used. the DREAMS subjects dataset includes whole-night PSG recordings of 20 healthy subjects and manually scored in sleep stages according to both the R&K and AASM criteria. Each recording includes at least two EOG channels, three EEG channels and one submental EMG channel recorded in sampling frequency of 200Hz. In this study one of the EOG channel and one re-referenced EEG channel were used after resampling to 100Hz. To be similar as much as possible to the other used datasets, Fp2-Cz EEG channel was acquired by re-referencing two EEG channels of Fp2-A1 and Cz-A1 in the recordings. This dataset has been used in this study to compare the performance of the proposed method to other studies. Therefore, LOO validation method was used to evaluate the performance same as EDF-20 dataset. Table 2 shows the average number and variation of epochs in each of the sleep stages for the train and test sets.

*Table 2. The number of 30s epochs in each validation approach. The values are presented in average±std style.*

|       |              | Splitting    | Wake           | S1             | S2             | SWS            | REM            |
|-------|--------------|--------------|----------------|----------------|----------------|----------------|----------------|
| Train | EDF Expanded | Night-out    | 223867±430.7   | 17058.4±675.8  | 54084.8±334.8  | 10266.4±432.7  | 20408.8±615.4  |
|       |              | Subject-out  | 223867±1024.9  | 17058.4±894.6  | 54084.8±455.4  | 10266.4±412.8  | 20408.8±379.4  |
|       |              | record-out   | 251695         | 18909          | 60258          | 11886          | 22674          |
|       | EDF-20       | LOO          | 36436.3±148.4  | 2060.6±62.3    | 8982.3±117.9   | 1969.4±67.4    | 2927.9±65.2    |
|       | DREAMS       | LOO          | 3402±88        | 1406±25.8      | 7838.5±100.7   | 3737.3±66.5    | 2868.1±34.2    |
| Test  | EDF Expanded | night-out    | 55927.8±432.3  | 4264.6±675.8   | 13521.2±334.8  | 2566.6±432.7   | 5102.2±615.4   |
|       |              | subject-out  | 55927.8±1022.7 | 4264.6±894.6   | 13521.2±455.4  | 2566.6±412.8   | 5102.2±379.4   |
|       |              | record-out   | 27925          | 2283           | 7313           | 987            | 2620           |
|       | EDF-20       | LOO          | 1917.7±148.4   | 108.5±62.3     | 472.8±117.5    | 103.7±67.4     | 154±65.2       |
|       | DREAMS       | LOO          | 179.1±88       | 74±25.8        | 412.6±100.8    | 196.7±66.5     | 151±34.2       |

### 6.2. Sleep Stage Scoring

First, using EDF-Expanded dataset, the single- and multi-epoch networks were trained by training sets in the night-holdout and subject-holdout approaches. After the training, the whole system, including both networks, was tested 5 times for each fold separately. The confusion matrices and results of the evaluation of the system for both approaches are presented in Tables 3 and 4.

Table 3. Confusion matrix and the result of evaluating the proposed method using the night-holdout validation approach (EDF-Expanded dataset).

|  |  | Multi-epoch network ($O^*_n$) | | | | | Recall |
|---|---|---|---|---|---|---|---|
|  |  | Wake | S1 | S2 | SWS | REM |  |
| Scores | Wake | 64.84±0.51 | 3.02±0.32 | 0.26±0.12 | 0.02±0.01 | 0.49±0.52 | 94.48±1.01 |
|  | S1 | 0.06±0.02 | 4.89±1.05 | 0.35±0.05 | 0.00±0.00 | 0.14±0.10 | 89.42±4.09 |
|  | S2 | 0.06±0.06 | 1.74±0.42 | 13.73±0.40 | 0.74±0.30 | 0.46±0.10 | 82.08±2.46 |
|  | SWS | 0.00±0.00 | 0.00±0.00 | 0.06±0.04 | 2.97±0.56 | 0.00±0.00 | 98.09±0.89 |
|  | REM | 0.01±0.01 | 0.11±0.02 | 0.15±0.04 | 0.00±0.00 | 5.90±0.85 | 95.78±0.46 |
|  | Precision | 99.80±0.13 | 49.81±4.31 | 94.35±1.32 | 80.08±3.46 | 85.24±5.88 |  |
|  | F1 | 97.07±0.57 | 63.95±4.52 | 87.77±1.26 | 88.16±2.43 | 90.12±3.45 |  |
|  | Accuracy = 92.33±0.59 | | Kappa = 0.85±0.00 | | MF1 = 85.41±1.57 | | |

Table 4. Confusion matrix and the result of evaluating proposed method using the subject-holdout validation approach (EDF-Expanded dataset).

|  |  | Multi-epoch network ($O^*_n$) | | | | | Recall |
|---|---|---|---|---|---|---|---|
|  |  | Wake | S1 | S2 | SWS | REM |  |
| Scores | Wake | 64.01±1.05 | 3.47±0.59 | 0.38±0.30 | 0.06±0.06 | 0.79±0.58 | 93.15±1.22 |
|  | S1 | 0.09±0.02 | 4.43±1.16 | 0.49±0.11 | 0.01±0.00 | 0.23±0.05 | 84.17±2.68 |
|  | S2 | 0.06±0.04 | 2.03±0.21 | 12.70±0.76 | 1.19±0.43 | 0.64±0.20 | 76.42±3.82 |
|  | SWS | 0.00±0.00 | 0.00±0.00 | 0.08±0.01 | 3.07±0.57 | 0.00±0.00 | 97.32±0.68 |
|  | REM | 0.01±0.01 | 0.19±0.04 | 0.19±0.12 | 0.00±0.00 | 5.88±0.54 | 93.71±1.68 |
|  | Precision | 99.74±0.08 | 43.38±4.80 | 91.86±2.40 | 71.00±9.37 | 78.45±7.12 |  |
|  | F1 | 96.33±0.62 | 57.18±4.70 | 83.40±2.87 | 81.86±6.22 | 85.30±4.78 |  |
|  | Accuracy = 90.08±1.63 | | Kappa = 0.81±0.03 | | MF1 = 80.81±3.15 | | |

The element $(i,j)$ of confusion matrices shows the rate (%) of the number of samples (30s epochs) which belong to sleep stage $i$ and were detected as sleep stage $j$ concerning the total number of samples in the test sets. Both recall and precision can be used to evaluate the performance of the system in each sleep stage separately. In both night-holdout and subject-holdout approaches, the proposed method had the best detection performance (recall) for the SWS stage and the worst for the S2 sleep stage. Also, in both set selection approaches, the proposed method has the highest retrieving rate (precision) and F1-score for the wake stage. As it is described, a probability-based sample section method was employed to prevent biasing the training process to specific sleep stages that had more samples (e.g., wake stage). Close recall values for different stages showed that the section method accomplished its goal. The proposed architecture achieved an average accuracy of 92.33%, Kappa of 0.85 and MF1 of 85.41% in night-holdout cross validation approach and average accuracy of 90.08%, Kappa of 0.81, and MF1 of 80.81% in the subject-holdout cross validation approach in EDF-Expanded dataset.

The results of the sleep stage scoring using EDF-Expanded dataset with just the single-epoch networks are presented in Table S1 and S2. The single-epoch network had an average Kappa of 0.75 and 0.72 for night-holdout and subject-holdout, respectively. Using the multi-epoch network to consider surrounding epochs increased the classification performance significantly and about 0.1 in Kappa score. Figure 4 shows an example of sleep scoring using the trained single- and multi-epoch networks.

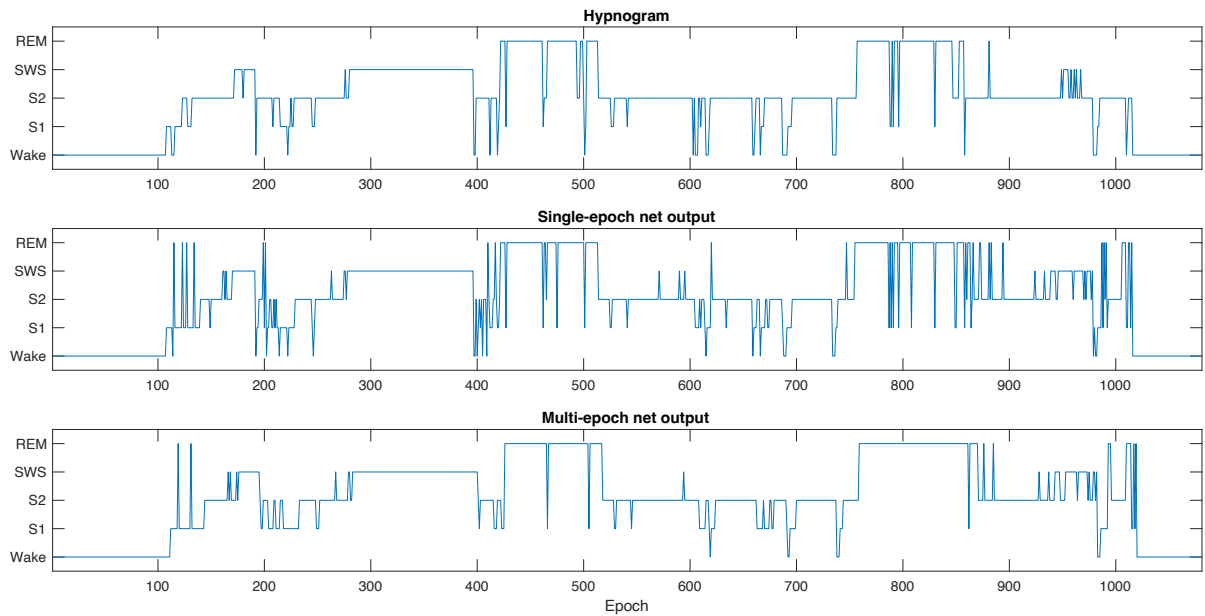

*Figure 4. An example of hypnogram (expert's scoring) and outputs of single- and multi-epoch networks for one of the recordings.*

By Comparing the output of single- and multi-epoch networks in Figure 4 we showed that the LSTM layers and considering the surrounding epochs caused the system to ignore single sudden changes. Table 5 shows the average number of epochs in which the single- and multi-epoch networks disagreed and how many samples were corrected by using the multi-epoch network.

Table 5. Confusion matrix of single- and multi-epoch networks for both validation approaches. The black numbers indicate the values of the confusion matrix. Green numbers show the number of samples in which the single-epoch network detected stages wrongly and the multi-epoch network corrected them. Red numbers show the number of samples which were scored correctly by the single-epoch network and the multi-epoch network changed the output wrongly (EDF-Expanded dataset).

|  |  | *Subject-holdout* | | | | |  | *Night-holdout* | | | | |
|---|---|---|---|---|---|---|---|---|---|---|---|---|
|  |  | Multi-epoch network ($O^*_n$) | | | | |  | Multi-epoch network ($O^*_n$) | | | | |
|  |  | Wake | S1 | S2 | SWS | REM |  | Wake | S1 | S2 | SWS | REM |
| Single-epoch network ($O_n$) | Wake |  | 488.2 / 62.0 / 392.0 | 105.2 / 44.8 / 55.5 | 3.2 / 0.0 / 2.8 | 110.8 / 56.8 / 43.0 |  |  | 309.2 / 42.2 / 246.0 | 100.0 / 70.8 / 23.8 | 0.8 / 0.0 / 0.2 | 71.8 / 37.2 / 28.8 |
|  | S1 | 1380.8 / 1343.8 / 32.0 |  | 673.2 / 523.2 / 91.2 | 0 | 571.2 / 366.5 / 75.8 |  | 1198.5 / 1171.5 / 23.5 |  | 914.5 / 774.2 / 91.5 | 0.2 / 0.2 / 0.0 | 424.8 / 273.5 / 59.2 |
|  | S2 | 223.8 / 221.2 / 2.0 | 632.0 / 245.5 / 335.8 |  | 5.8 / 0.5 / 5.2 | 244.5 / 154.8 / 74.0 |  | 143.0 / 140.0 / 2.8 | 433.0 / 169.8 / 236.8 |  | 4.0 / 0.8 / 3.2 | 200.2 / 121.0 / 66.8 |
|  | SWS | 132.5 / 131.0 / 0.2 | 11.8 / 3.0 / 1.0 | 561.0 / 487.0 / 46.8 |  | 4.5 / 2.5 / 0.5 |  | 68.0 / 65.5 / 1.0 | 5.0 / 2.2 / 0.0 | 471.2 / 432.8 / 30.2 |  | 1.8 / 1.2 / 0.0 |
|  | REM | 1300.2 / 1290.5 / 2.2 | 449.8 / 152.5 / 24.2 | 504.2 / 394.2 / 36.5 | 1.5 / 0.0 / 0.0 |  |  | 816.8 / 803.8 / 2.8 | 260.0 / 70.0 / 12.0 | 531.0 / 438.5 / 29.2 | 0.2 / 0.0 / 0.0 |  |

The confusion matrices values in Table 5 show that using the multi-epoch network on top of the single-epoch network outputs improved staging performance in most situations. Generally, as it can be inferred by comparing overall scores, using the multi-epoch network improve the performance of the system, and in most disagree cases the multi-epoch network corrected the detection. The pair of S1-Wake and REM-Wake ($O_n$ -$O^*_n$) had the greatest number of disagreed outputs and in more than 95.5% of cases, the multi-epoch network corrected the detection. On the other hand, the worst cases were pairs of Wake-S1 and S2-S1 where the multi-epoch network changed the output to S1 wrongly in 80% and ~53.5% of disagreed samples.

The proposed architecture was trained using the EDF-20 and DREAMS datasets and evaluated by LOO cross validation method separately. The proposed method achieved accuracy of 93.9%, Kappa of 0.88 and MF1 of 83.34% for the EDF-20 dataset (The confusion matrix of the results for the EDF-20 dataset is presented in Table S3). By using the DREAMS dataset, after training and evaluating the proposed architecture, the network achieved accuracy of 88.09%, Kappa of 0.84 and MF1 of 86.96% in LOO cross validation (The confusion matrix of the results for the DREAMS dataset is presented in Table S4).

### 6.2.1. Comparison

Table 6 shows the overall performance comparison of the proposed method and some recent deep learning studies that used the Physionet EDF datasets including EDF-Expanded and EDF-20.

*Table 6. Result comparison between the proposed method and some other deep learning studies that used the EDF-Expanded dataset or a subset of that.*

| Method | Records | Channels | Validation | Acc | MF1 | Kappa |
|---|---|---|---|---|---|---|
| stacked sparse autoencoders [101] | 20 Subjects | EEG | LOO | 78.9 | 73.7 | - |
| DeepSleepNet [48] | 20 Subjects | EEG | LOO | 82.0 | 76.9 | 0.760 |
| Personalized Deep CNN [102] | 19 Subjects | EEG-EOG | 2$^{nd}$ Night-out | 84.0 | - | - |
| Attentional RNN [65] | 20 Subjects | EEG | LOO | 79.1 | 69.8 | 0.700 |
| SleepEEGNet [103] | 20 Subjects | EEG | 20-fold CV | 84.3 | 79.66 | 0.79 |
| Multitask 1-max CNN [104] | 20 Subjects | EEG-EOG | LOO | 82.3 | - | 0.74 |
| SeqSleepNet+fine tuning [105] | 20 Subjects | EEG-EOG | 10-fold CV | 84.3 | 77.7 | 0.776 |
| DeepSleepNet+fine tuning [105] | 20 Subjects | EEG-EOG | 10-fold CV | 84.6 | 79.0 | 0.782 |
| CWT and transfer Learning [106] | 61 Records | EEG | 45 training, 10 testing | 85.1 | - | - |
| SeqSleepNet+regularization [107] | 75 Subjects | EEG | 2$^{nd}$ Night-out | 82.6 | 76.4 | 0.760 |
| Scattering spectrum diffusion [108] | 20 Subjects | EEG | LOO | 84.44 | 78.25 | 0.784 |
| Residual Attention Model [109] | 20 Subjects | EEG | LOO | 84.3 | 79.0 | 0.78 |
| Graph-Temporal fused CNN [110] | 20 Subjects | EEG | 10-fold CV | 89.1 | - | 0.829 |
| *The proposed method* | 75 Subjects | EEG-EOG | 15 subject-out | **90.08** | **80.81** | 0.81 |
| *The proposed method* | 75 Subjects | EEG-EOG | 30 nights-out | **92.33** | **85.41** | **0.85** |
| *The proposed method* | 20 Subjects | EEG-EOG | LOO | **93.94** | **83.34** | **0.88** |
| *The proposed method* | 20 Subjects | EEG | LOO | **92.65** | **81.88** | **0.85** |

The majority of prior studies (Table 6) deployed a cross-validation approach in which the ratio of test recordings to all recordings is less than or equal to 0.1. In the current study, for EDF-Expanded dataset this ratio is about 0.2. This means the proposed method has been tested in more difficult conditions and achieved the best overall performances in comparison to Table 6 presented studies, except for the Kappa measure in the subject-holdout validation approach. The Kappa of [110] is higher than our method in the subject-holdout validation approach but that study did not separate the recordings into training and test processes and the trained network was fed by a subset of all recording epochs. Also, we used both EEG and EOG signals in the proposed system, which could increase the overall performance, especially in S1 and REM detection as there are eye movements in these two stages which can be reflected on the EOG signal. Despite that, our methods still show higher performance in comparison with other studies that used both EEG and EOG. To increase results comparability, the EDF-20 dataset was used which includes 20 records and evaluated using LOO cross validation method same as many other mentioned studies in Table 6. Also, as many studies have tried to proposed sleep stage scoring system using just single EEG channel data, we repeated the experiments, trained and evaluated the proposed method by removing the EOG Gabor layer and used a single EEG channel. The results of using EDF-20 dataset in both condition of using EEG and EOG as well as single EEG channel are proposed in Table 6 (The confusion matrix of the results for using single channel EEG of the EDF-20 dataset is presented in Table S5). In the both conditions, the proposed method over performed in comparison to all other studies in Table 6.

The same experiments of using both EEG-EOG signals and just EEG signals were repeated using the DREAMS dataset (The confusion matrices of the both experiments are presented in Table S4 and S6). Table 7 shows the results of the proposed method using DREAMS dataset in comparison to some other studies which used the same dataset.

*Table 7. Result comparison between the proposed method and some other studies that used the DREAMS dataset.*

| Method | Channels | Acc | MF1 | Kappa |
|---|---|---|---|---|
| AT-BiLSTM [119] | EEG | 81.72 | 80.74 | 0.75 |
| HMM+Random Forest [120] | EEG | 80.35 | - | 0.73 |
| CNN-CRF [121] | EEG | 83.3 | 76.7 | 0.76 |
| Bagged tree [122] | EEG | 79.9 | - | - |
| SBLE+SVM [117] | EEG | 83.3 | - | 0.77 |
| XGBoost [123] | EEG | 85.8 | - | - |
| *The proposed method* | EEG-EOG | **88.09** | **86.96** | **0.84** |
| *The proposed method* | EEG | **86.71** | **85.67** | **0.82** |

The results in Table 7 shows that the proposed method in the both condition of using EEG and EOG, and single EEG channel over performed in comparison to the other mentioned studies.

Stage S1 is the most difficult stage to detect [111] with the lowest agreement between different scorers [18]. Table 8 shows a comparison between the proposed method's performance in stage S1 and some other studies without considering the main method (shallow learning or deep learning) which proposed performance for each sleep stage separately.

*Table 8. Comparison of performance of stage S1 detection between the proposed method and some other studies which used the same dataset or a subset of that. (The [+] mark indicate studies with shallow classifier model)*

|  | S1 detection performance | |
|---|---|---|
| **Study** | **Recall (%)** | **F1-score (%)** |
| Liang et al. [112][+] | 18.7 | |
| Liang et al. [111][+] | 30 | - |
| Hsu et al. [113][+] | 36.70 | - |
| Zhu et al. [114][+] | 15.8 | - |
| Hassan et al. [115][+] | 37.42 | - |
| Liang et al. [116][+] | 39.6 | - |
| Tsinalis et al. [101] | - | 47.0 |
| Tsinalis et al. [47] | - | 43.7 |
| Supratak et al. [48] | 50.1 | 46.6 |
| Seifpour et al. [117][+] | 40.1 | - |
| Phan et al. [65] | - | 34.2 |
| Mousavi et al. [103] | 54.51 | 52.19 |
| Humayun et al. [118] | 39.2 | 48.7 |
| Phan et al. [104] | 31.9 | - |
| Phan et al. [105] | - | 50.0 |
| Hongzhe Li [55] | 44 | 44 |
| Jadhav et al. [106] | 36.11 | - |
| Phan et al. [107] | 45.7 | - |
| Liu et al. [108] | 46 | 43 |
| Qu et al. [109] | 47.6 | 48.3 |
| Cai et al. [110] | 33.11 | - |
| *The proposed method (EDF-Expanded, Night-holdout)* | **89.42** | **63.95** |
| *The proposed method (EDF-Expanded, Subject-holdout)* | **84.17** | **57.18** |
| *The proposed method (EDF-20, LOO)* | **89.27** | **67.46** |
| *The proposed method (EDF-20, LOO, just EEG)* | **85.16** | **65.26** |
| *The proposed method (DREAMS, LOO)* | **91.86** | **77.23** |
| *The proposed method (DREAMS, LOO, just EEG)* | **91.02** | **76.58** |

Table 8 shows the proposed method has significantly better S1 detection performance in both recall and F1-score measures.

In sum, the proposed system for automatic sleep stage scoring had good potential in learning the knowledge from the signals and showed a noticeable performance by evaluating in different ways. By using the Night-holdout and LOO evaluation approaches, we showed the proposed system had better performance and generalizability even in comparison to systems with lenient evaluation. In the following sections, we analyzed a trained system to demonstrate and interpret the learned knowledge by the system at different levels.

### 6.3. Interpretation

The proposed system was trained and tested using EDF-Expanded dataset and the record-holdout evaluation approach to increase the system's learned knowledge and interpretability by the user. The purpose of the training and evaluation was not to have an automatic sleep stage scoring system, but to have well-trained networks that could facilitate translation in the application. The overall performance of the trained system was an accuracy of 92.25%, MF1 of 84.1%, and Kappa of 0.85, demonstrating a well-trained system.

At the first level of interpretation, the optimized Gabor kernels of the trained single-epoch network represented the standard waveforms used for scoring. During the training process, the parameters of the kernels were optimized to find the best waveform, which after convolving to the EEG and EOG signals produce Gabor layer outputs ($GL^i(t)$). The convolution operation in the Gabor layer is similar to calculating the power of the kernel's waveform in the input signal. Therefore, the trained Gabor kernel represented the optimized waveforms for sleep staging. As the frequency domain in EEG signals is the most important characteristic of the waveform in sleep signals, the trained Gabor kernels and their frequency domain are presented in Figure S1 and S2. The optimized (trained) Gabor kernels covered several frequency ranges, some of which fitted to relevant sleep EEG waveforms. Figure 5 represents some of the optimized Gabor waveforms along with their frequency domain.

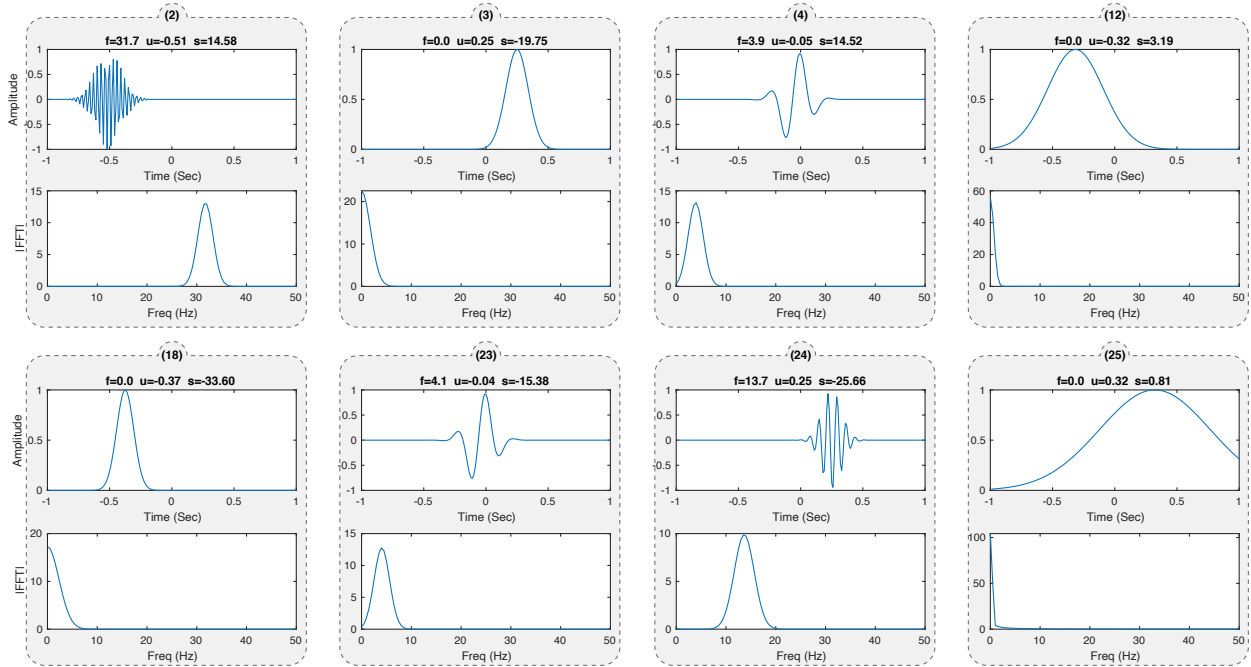

*Figure 5. Waveform and frequency domain of some optimized Gabor kernels.*

Figure 5 shows that the Gabor kernels covered a wide range of frequencies. Some of the optimized kernels were perfectly matched to well-known EEG waveforms including slow waves (SW, ~1Hz), theta waves, spindle. In Figure 5, optimized Gabor kernels 3 and 18 are similar to SW waveform and delta waves (0.5 to 3 Hz), 4 and 23 with the main frequency of ~4Hz are fitted to theta waves, and 24 is similar to a sleep spindle.

The optimized kernels do not represent the importance of each kernel and should be considered the amount that each kernel impacted the output of the single-epoch network. Therefore, $Eff_i$ and $\#_i$ quantifiers were used to represent the importance of the optimized Gabor kernels in both qualitative and quantitative manner. Figure 6-a and 6-b show the overall relative impact of each Gabor kernel on the sleep staging process of the single-epoch network. Base on Figure 6-a representation, different optimized Gabor kernels had different qualitative impacts on the decision-making process. For example, Gabor kernel 18 which is similar to SW and delta waves, and behaves as a lowpass filter had the most impact. The amplitude of delta waves corresponds to the amount of synchronization of cortical pyramidal cells and delta waves can represent a progression from waking to SWS [124] [78]. The time course of delta activity over the night is well described and decreases across the night [125], [126] [79], [80]. Therefore, we expected SW and delta waves to have the most leverage in the sleep staging process. On the other hand, the result in Figure 6-a show that some Gabor kernels were not important because the training process could not optimize them or they had redundant information, and other optimized kernels produced enough information for decision making. For example, Gabor kernels 6, 13, 16, 20, and 21 were not optimized well and did not have a considerable effect on the staging process. On the other hand, Gabor kernels 3, 7, 8, 10, 18, 19, 26, 27, and 28 were almost equal and with a waveform of SW behaved as lowpass filter but except for kernels 8, 10, 18 the others did not have a considerable effect on the staging process possibly due to redundancy of information.

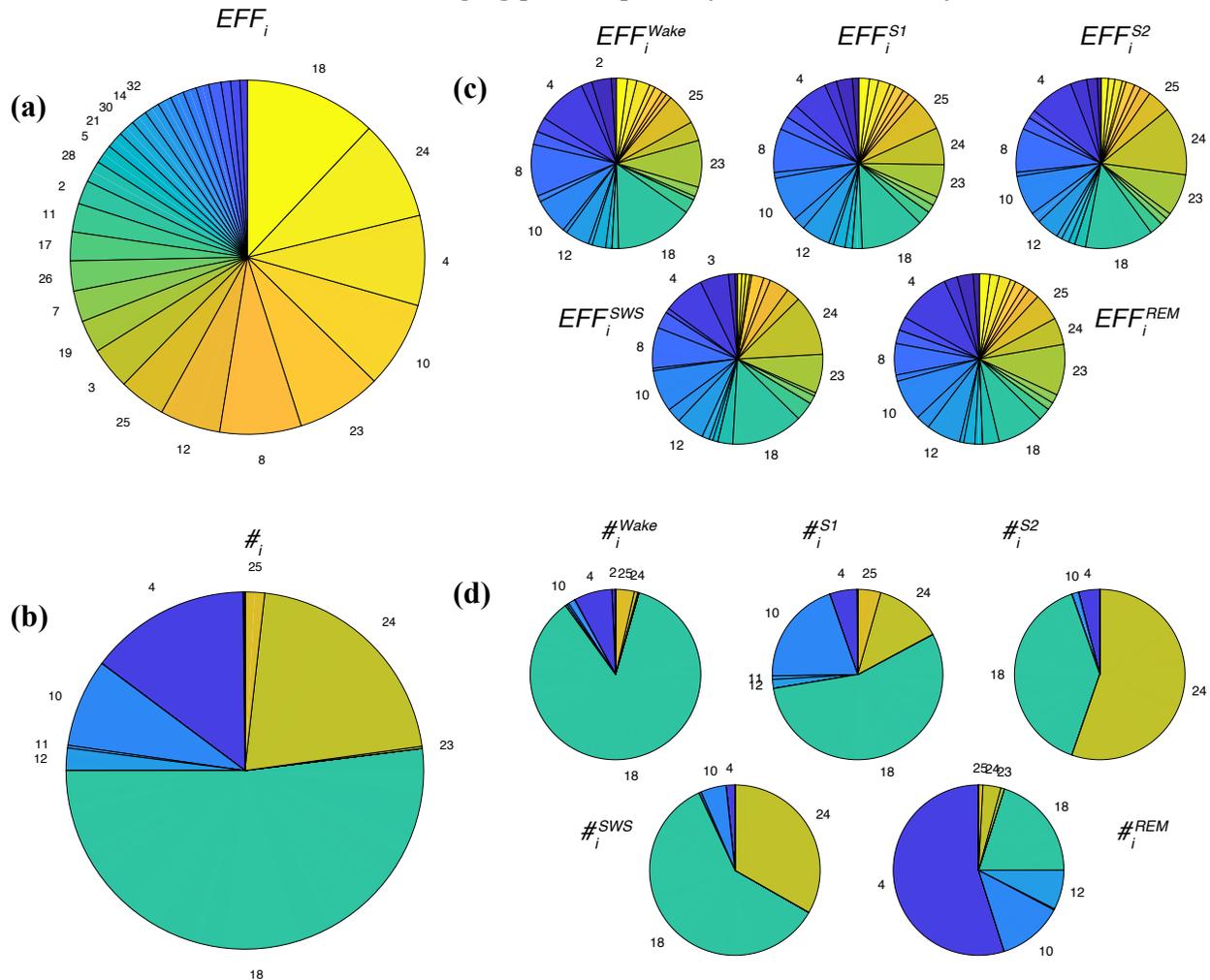

*Figure 6. Impact of the Gabor kernels on sleep stage scoring. a) overall qualitative impact of all optimized Gabor kernels on the sleep stage scoring process (*$Eff_i$ *quantifier). b) overall quantitative impact of all optimized Gabor kernels on the sleep stage scoring process (*$\#_i$ *quantifier). c) overall impact of all optimized Gabor kernels on each of the sleep stages separately (*$Eff_i^j$ *quantifier). d) The ratio of how many times each Gabor kernel had the highest impact on each of the sleep stages (*$\#_i^j$ *quantifier).*

Figure 6-b shows the ratio of how many times each Gabor kernel had highest impact on the overall sleep stage scoring process. As the $\#_i$ is a quantitative quantifier, it can represent which Gabor kernel affected the selected sleep stage greatly most of the time. Based on the results in Figure 6-b, kernel 18 and 10 as SW representors had the greatest number of having highest impact, kernel 24 as a sleep spindle representor had the next greatest number of highest impact and kernel 4 as a theta waveform (3.5 – 7.5 Hz) representor had the next greatest number of highest impact on sleep stage scoring process.

At the next level, we measured the quantity and quality of Gabor kernel waveforms in each stage separately using $Eff_i^j$ and $\#_i^j$ respectively. Figure 6-c represents the $Eff_i^j$ quantifier for all the kernels in different sleep stages. The results in this figure are compatible to the experts' knowledge and the sleep scoring manuals. For example, the kernels which represent SW and spindles have highest impact in stage S2 and SWS ($Eff_{[18,12,10,8],[24]}^{S2,SWS}$) and the value of $Eff_{24}^{S2}$ which represent the effect of spindle in stage S2 is greater than $Eff_{24}^{SWS}$ which represent the effect of spindle in stage SWS which is compatible to expert's knowledge as the sleep spindle is the main signature of stage S2 [127] and occurs less frequently during the SWS [128]. Similarly, the values of SW representors (kernels 8, 10, 12 and 18) in stage SWS ($Eff_{[8,10,12,18]}^{SWS}$) have greater sum values in comparison to stage S2 ($Eff_{[8,10,12,18]}^{SWS}$) which agree with experts' knowledge. Figure 6-d shows the relative times that each Gabor kernel had the highest impact on each sleep stage. It is important to consider the values for each stage in comparison with other stages. Generally, kernels that were related to SW (such as kernel 18) had a big effect on all the stages. But, the comparison of different stages showed for stage S1 kernel 25 which is a very slow wave representor, similar to slow eye-rolling and movement artifact on the EEG signals, played an important role. Spindle-like waveform (kernel 24) had the highest impact in sleep stage S1 more than half of the time. Another important difference between stages in Figure 6-d is kernel 4 in the REM stage. Kernel 4 is a theta waveform and during REM sleep cortical theta wave activity is predominant [129]. Interestingly, in more than half of REM epochs, the theta waveform (kernel 4) had the highest effect in REM stage detection.

Also, Figure 7 shows the variation range of the impacts for these kernels ($Eff_i^j$ quantifier) across stages and marks significant differences (using paired t-test, p-value < 0.05) between them in different stages.

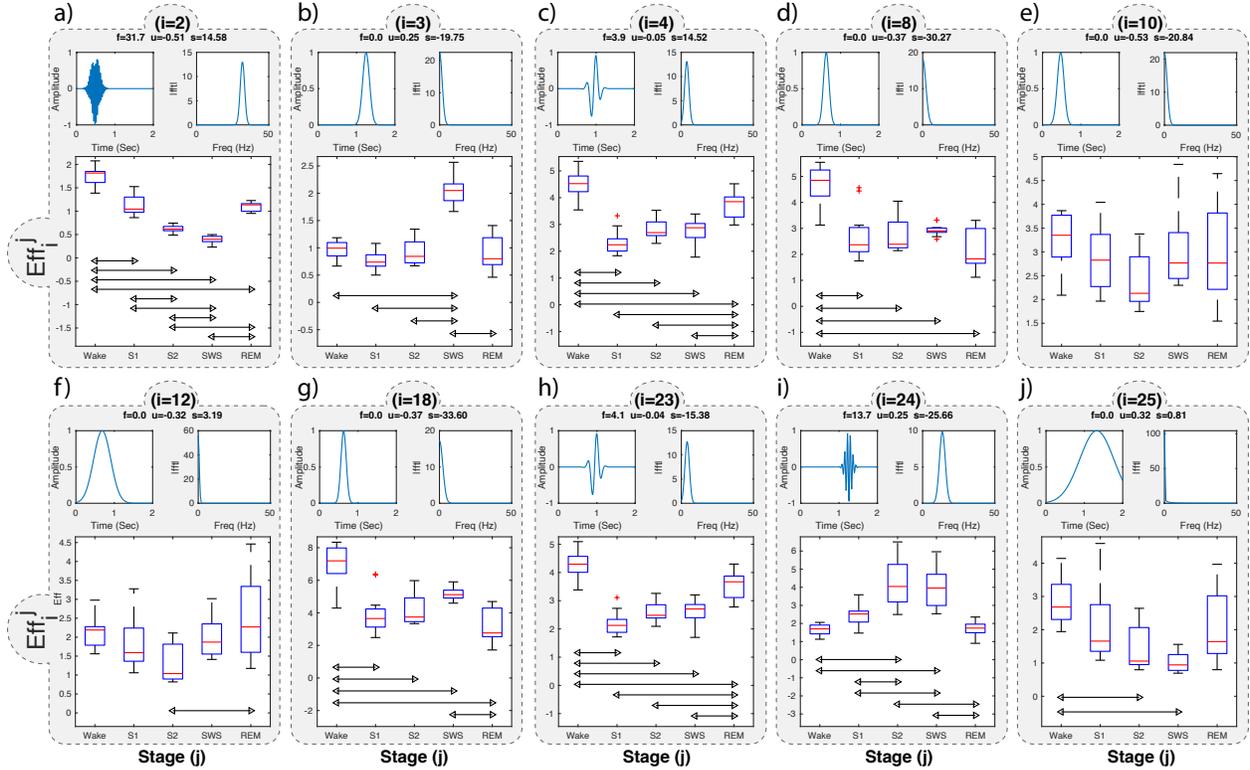

*Figure 7. The variation of $\overline{Eff}$ for some Gabor kernels in different stages and the result of the statistical tests. The significant differences (p-value<0.05) are indicated using bidirectional arrows.*

Kernel 2 (Figure 7-a) is a high frequency waveform that decreases from wake to SWS. Figure 7-b, d, e, and g shows the distributions for waveforms which were similar to SW. As it is mentioned, SW existence is important for S2 and SWS stage detection, and Figure 7-b and g represent SW's high impact on SWS. However, kernel 8 and 10 had redundant information for sleep stages. Kernel 4 and 23 (Figure 7-c and h) with the main frequency of 4 Hz acted as a bandpass filter in a frequency range around the delta and theta boundary. The statistical tests showed this waveform had the highest impact in REM in comparison to other NREM sleep stages. Figure 7-f and j the very low frequency can be interpreted as slow rolling eye movements and movement artifact on EEG signals and a result showed lowest impact on stage S2 and SWS stage detection. Kernel 24 (Figure 7-i) as is described above, as spindle waveform, had the highest impact on stage 2 and SWS which is consistent with previous knowledge. Concluding, the values in Figure 7 and the results of statistical tests showed the learned knowledge from the single-epoch network, and the interpretation of the effect of each optimized kernel waveform were compatible with scorer knowledge.

In the next level of interpretation, based on the kernel impacts on outputs, for each 30s epoch, we determined how much EEG and EOG signal impacted the output by averaging the effect quantifier for EEG related kernels ($Eff_a$ where $a \in [EEG\ kernels]$) and EOG related kernels ($Eff_b$ where $b \in [EOG\ kernels]$). Therefore, the ratio of EEG to EOG impact was defined and Figure 8 shows these ratio values for different epochs in different stages.

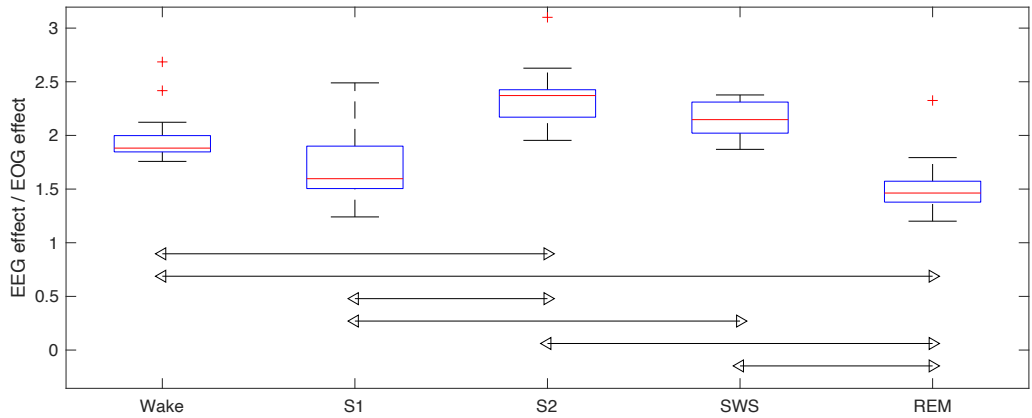

*Figure 8. The variation of EEG/EOG impact ratio for different stages. The significant differences (p-value<0.05) are indicated using bidirectional arrows.*

Given that eyes movement is one the main characteristics of sleep stage S1 and REM and scorers used EOG signal to determine eye movement [14], [15] we expected to show a dependency of sleep stages S1 and REM on EOG signals. Values in Figure 8 and the result of the statistical test showed the ratio of EEG/EOG impact on REM and S1 stages were significantly lower than other sleep stages. Also, there was no significant difference between stages S2 and SWS with respect to the ratio of EEG to EOG impact which was compatible with the prior knowledge as there is no eye movement in these to sleep stages. Comparing single- and multi-epoch networks outputs, the system learned when it became necessary to consider surrounding epochs.

In the next level, as the most detailed level, we interpreted the impact of each Gabor kernel on the output over time by using $Eff(t)$ time series by considering both amplitude of the kernel output (amplitude of the kernel waveform in the input signal) and positive sensitivity of the kernel output. Therefore, by using $Eff(t)$, we showed which waveforms had the highest impact on the output of the single-epoch network over the time. Figure 9 shows the $Eff(t)$ time series for three different EEG epochs in different stages for the two Gabor kernels with the highest impact.

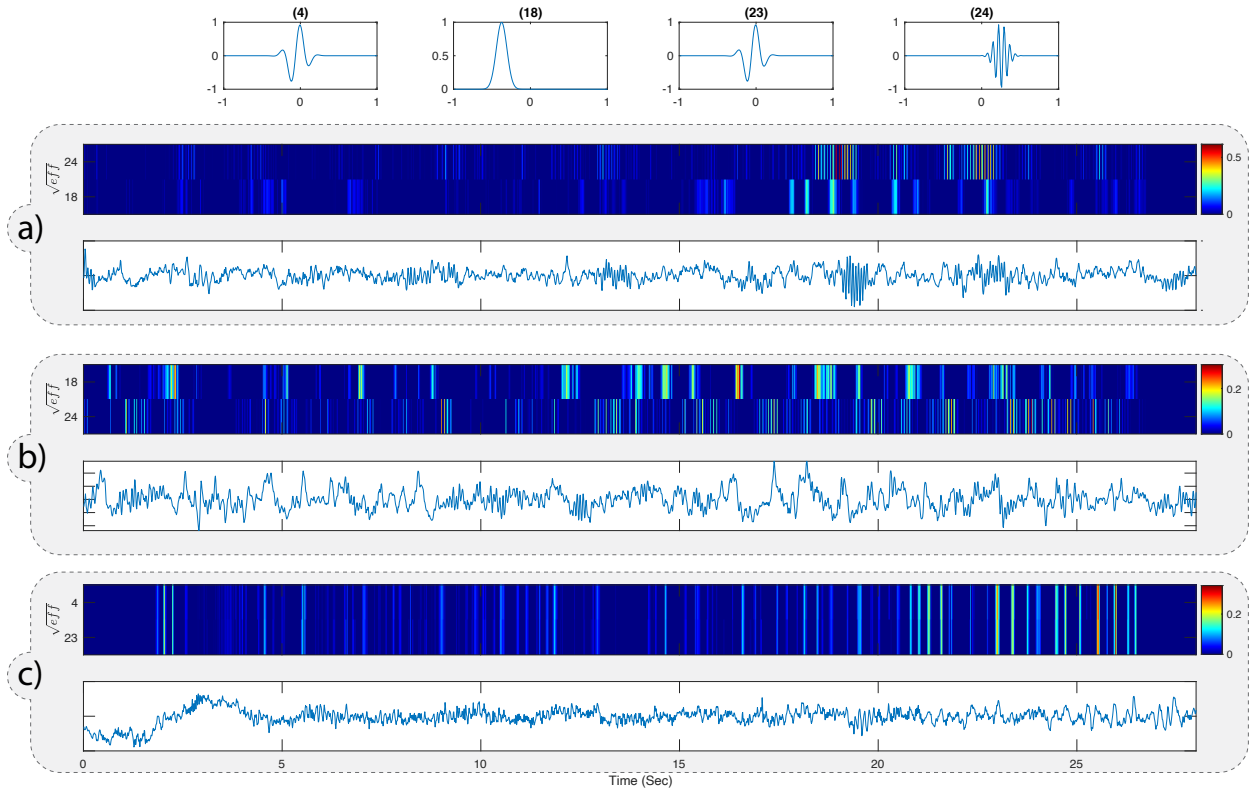

*Figure 9. Some samples of $Eff(t)$ time series for two Gabor kernels with the highest impact along with related EEG signal. a) A sample of EEG signal for sleep stage S2. b) sleep stage SWS. c) Sleep stage REM.*

Figure 9 show three examples of the proposed method for interpretability. In Figure 9-a, the input epoch is S2 and shows that spindles and SOs play important roles in the detection of this stage. $Eff(t)$ time series in Figure 9-a and b could represent when and the magnitude that each spindle or SW had on stage detection. For example, in Figure 9-a, the spindle just before second 20 and right around second 23 had the highest impact in S2 stage detection, and in Figure 9-b at around 10 seconds, SW and spindles played more important roles in stage detection. In Figure 9-c, theta-like waveforms (kernels 4 and 23) were detected and had the highest impact in the last 10 seconds of the EEG epoch.

### 6.4. Effect of Gabor Layer

The effect of the Gabor layers in the proposed method could be studied in two different categories of sleep stage scoring performance and interpretability. To study the effect of the Gabor layers on the proposed method, the Gabor layers replaced by one-dimensional convolution layers with the same length of 200 (2 seconds with sampling rate of 100 Hz) then the networks were trained and evaluated with the LOO cross validation using EDF-20 dataset. The trained networks (without Gabor layers) achieved accuracy of 84% and Kappa of 0.77 which were much less performed than the networks with Gabor layers. Also, the train process of the networks with and without Gabor layers were captured to make a comparison. Figure 10 shows the loss and Kappa of training and validation sets during the training process of two single-epoch networks with and without Gabor layers.

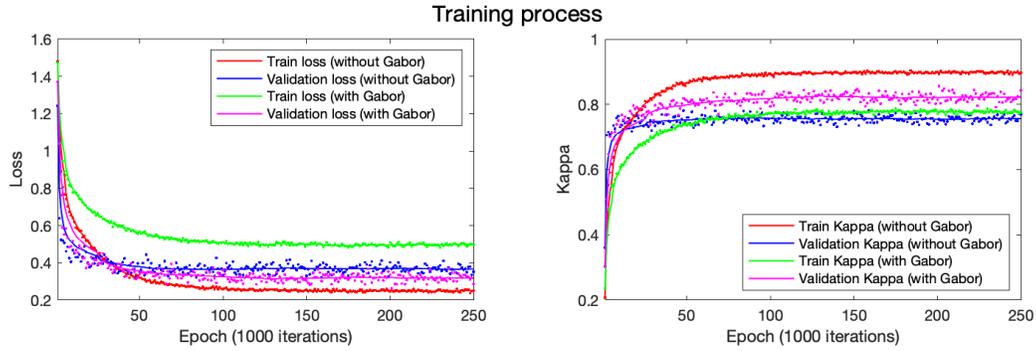

*Figure 10. Training process of single-epoch network with and without the Gabor layers. The left plot shows the loss of the training and validation set of the data during the training process and the right plot shows the Kappa measure in the same process.*

Results of measuring the loss and Kappa during the training process in Figure 10 shows that although the network without the Gabor layers learned faster in the first epochs but it was overfitted faster and the network with Gabor layers was more generalizable as the validation losses were always less than train losses and the Kappa values were greater in the validation data set. The network without Gabor layers was overfitted after about 30 epochs (30000 iterations) while the loss was decreasing and the Kappa was stabled. As a conclusion in comparison between sleep stage scoring performance between networks with and without the Gabor layers, the networks without the Gabor layers had less generalizability and weaker overall performance in comparison to the networks with the Gabor layers.

After training the networks without the Gabor layers, the first one-dimensional convolution layer can represent information similar to the Gabor layers in the main proposed network. However, as Figure S4 represent EEG signal related filters in the first one-dimensional convolution layer, the filters waveform would not contain any specific interpretable information of the waveforms from the EEG signal in both time and frequency domain. Therefore, by replacing the Gabor kernels with one-dimensional filters in the first convolution layer, almost all interpretability that the proposed architecture could provide would be wasted. The only remain interpretability is using some sensitivity based methods, such as saliency map [86], and Grad-CAM [89], which can provide information about which part of the input signals affected the output more without considering any specific waveforms or patterns that the experts use during the manual sleep stage scoring.

## 7. Conclusion

We proposed an interpretable DL system for sleep stage scoring by using prior knowledge about the standard waveform characteristics used in the scoring process, as well as considering surrounding epochs, and EOG signals to detect some specific stages. The proposed method leaves open the possibility for adding more explanatory signals that contain standard features, such as ECG. A new convolutional-based layer was defined by using the Gabor function as the kernel. Different cross validation approaches and datasets were considered to evaluate the performance of the proposed method and its generalizability. The results showed superior performance of the proposed method in comparison to prior work with even more rigorous evaluation conditions in the current model.

Using Gabor functions and optimizing them provided levels of interpretability across different aspects. The optimized Gabor kernels were similar to some of the well-known EEG waveforms. Based on different interpretation results it was shown that the learned knowledge by the proposed system from the data was compatible with prior knowledge which the scorers used for the stage scoring process. Also, we showed effect of using the Gabor layer could improve both performance and interpretability in comparison to same architecture without Gabor layer. This study presented the idea of using bounded kernels in a convolutional layer to design an interpretable system for analyzing the signals which include some specific waveforms such as EEG signals. However, in the proposed impact quantification we just used the positive gradients and considered the waveforms which existed in the signals. But it is possible to consider the negative gradients to quantify the oppositional impact of different kernels on outputs and to show the impact of waveforms which do not exist in the input signals which can be a potential approach for future studies.

# Supplementary

*Table S. 1. Confusion matrix and result of evaluation the single-epoch network using the night-holdout validation approach (EDF-Expanded dataset).*

|  |  | Single-epoch network ($O_n$) | | | | | Recall |
|---|---|---|---|---|---|---|---|
|  |  | Wake | S1 | S2 | SWS | REM |  |
| Scores | Wake | 62.54±1.36 | 4.10±0.76 | 0.34±0.13 | 0.10±0.06 | 1.55±0.88 | 91.12±2.23 |
|  | S1 | 0.09±0.01 | 4.75±1.05 | 0.42±0.03 | 0.01±0.01 | 0.17±0.10 | 86.88±4.30 |
|  | S2 | 0.16±0.09 | 2.34±0.83 | 11.99±0.80 | 1.27±0.16 | 0.96±0.25 | 71.71±4.21 |
|  | SWS | 0.00±0.00 | 0.00±0.00 | 0.02±0.03 | 3.01±0.57 | 0.00±0.00 | 99.30±0.70 |
|  | REM | 0.05±0.03 | 0.44±0.10 | 0.25±0.09 | 0.01±0.01 | 5.42±0.76 | 88.05±1.48 |
|  | Precision | 99.53±0.20 | 40.82±5.24 | 92.11±1.81 | 68.29±1.32 | 67.84±6.73 |  |
|  | F1 | 95.13±1.31 | 55.46±5.47 | 80.57±2.54 | 80.92±0.82 | 76.51±4.74 |  |
|  | **Accuracy** = 87.72±1.96 | | **Kappa** = 0.75±0.02 | | **MF1** = 77.72±2.30 | | |

*Table S. 2. Confusion matrix and result of evaluation the single-epoch network using the subject-holdout validation approach (EDF-Expanded dataset).*

|  |  | Single-epoch network ($O_n$) | | | | | Recall |
|---|---|---|---|---|---|---|---|
|  |  | Wake | S1 | S2 | SWS | REM |  |
| Scores | Wake | 60.83±2.59 | 4.66±0.94 | 0.51±0.25 | 0.22±0.13 | 2.50±1.46 | 88.51±3.62 |
|  | S1 | 0.15±0.05 | 4.07±1.17 | 0.63±0.18 | 0.02±0.02 | 0.37±0.13 | 77.15±6.62 |
|  | S2 | 0.18±0.08 | 2.10±0.15 | 11.40±0.96 | 1.78±0.47 | 1.15±0.42 | 68.63±4.82 |
|  | SWS | 0.00±0.00 | 0.00±0.00 | 0.02±0.01 | 3.13±0.57 | 0.00±0.00 | 99.20±0.47 |
|  | REM | 0.09±0.02 | 0.60±0.08 | 0.32±0.10 | 0.01±0.01 | 5.26±0.55 | 83.75±3.07 |
|  | Precision | 99.32±0.22 | 35.30±6.62 | 88.52±3.34 | 60.76±8.46 | 58.43±12.21 |  |
|  | F1 | 93.57±2.09 | 48.31±7.32 | 77.28±4.06 | 75.11±6.46 | 68.37±9.15 |  |
|  | **Accuracy** = 84.69±3.56 | | **Kappa** = 0.72±0.06 | | **MF1** = 72.53±4.97 | | |

Table S. 3 Confusion matrix and the result of evaluating proposed method using LOO validation approach for EDF-20 dataset.

|  |  | Multi-epoch network ($O^*_n$) | | | | | Recall |
|---|---|---|---|---|---|---|---|
|  |  | **Wake** | **S1** | **S2** | **SWS** | **REM** |  |
| **Scores** | **Wake** | 66.61±4.98 | 1.46±0.92 | 0.13±0.16 | 0.04±0.11 | 0.74±1.30 | 96.62±2.48 |
|  | **S1** | 0.04±0.07 | 3.36±1.96 | 0.25±0.28 | 0.00±0.01 | 0.09±0.09 | 89.27±5.87 |
|  | **S2** | 0.02±0.04 | 1.20±1.27 | 14.48±3.93 | 1.09±1.33 | 0.61±0.67 | 83.66±8.76 |
|  | **SWS** | 0.00±0.00 | 0.00±0.01 | 0.15±0.13 | 3.85±2.39 | 0.00±0.02 | 95.05±4.83 |
|  | **REM** | 0.01±0.03 | 0.07±0.07 | 0.15±0.28 | 0.00±0.00 | 5.64±2.45 | 91.03±21.96 |
|  | **Precision** | 99.90±0.17 | 55.33±12.64 | 95.25±3.73 | 73.01±24.79 | 80.54±17.78 |  |
|  | **F1** | 98.22±1.30 | 67.46±9.38 | 88.85±5.88 | 80.21±19.59 | 81.97±21.18 |  |
|  | **Accuracy** = 93.94±2.52 | | **Kappa** = 0.88±0.06 | | **MF1** = 83.34±6.57 | | |

Table S. 4 Confusion matrix and the result of evaluating proposed method using LOO validation approach for DREAMS dataset.

|  |  | Multi-epoch network ($O^*_n$) | | | | | Recall |
|---|---|---|---|---|---|---|---|
|  |  | **Wake** | **S1** | **S2** | **SWS** | **REM** |  |
| **Scores** | **Wake** | 14.43±9.44 | 0.55±0.59 | 0.03±0.06 | 0.02±0.04 | 0.08±0.17 | 94.00±7.74 |
|  | **S1** | 0.17±0.17 | 6.94±2.82 | 0.30±0.42 | 0.01±0.02 | 0.13±0.16 | 91.86±7.22 |
|  | **S2** | 0.32±0.28 | 2.33±1.52 | 33.79±8.52 | 3.99±2.27 | 1.38±1.01 | 80.30±5.59 |
|  | **SWS** | 0.02±0.05 | 0.01±0.03 | 1.87±2.30 | 18.23±6.06 | 0.00±0.00 | 90.88±8.57 |
|  | **REM** | 0.07±0.13 | 0.20±0.29 | 0.43±0.81 | 0.01±0.02 | 14.69±3.33 | 95.64±5.34 |
|  | **Precision** | 95.67±2.97 | 68.43±14.54 | 92.76±6.02 | 80.66±15.31 | 90.41±5.42 |  |
|  | **F1** | 94.69±5.09 | 77.23±10.42 | 85.87±4.16 | 84.19±10.63 | 92.80±3.94 |  |
|  | **Accuracy** = 88.09±2.54 | | **Kappa** = 0.84±0.04 | | **MF1** = 86.96±3.32 | | |

Table S. 5 Confusion matrix and the result of evaluating proposed method using LOO validation approach for EDF-20 dataset using single EEG channel signals.

|  | Multi-epoch network ($O^*_n$) | | | | | Recall |
|---|---|---|---|---|---|---|
|  | **Wake** | **S1** | **S2** | **SWS** | **REM** |  |

|  | Wake | S1 | S2 | SWS | REM | Recall |
|---|---|---|---|---|---|---|
| **Wake** | 65.36±5.13 | 1.76±1.51 | 0.62±1.46 | 0.05±0.11 | 1.19±1.81 | 94.90±5.07 |
| **S1** | 0.05±0.06 | 3.25±2.02 | 0.27±0.33 | 0.00±0.01 | 0.17±0.22 | 85.16±11.45 |
| **S2** | 0.11±0.37 | 0.87±1.00 | 14.56±4.12 | 1.33±1.40 | 0.53±0.57 | 84.03±9.37 |
| **SWS** | 0.00±0.01 | 0.00±0.00 | 0.13±0.10 | 3.87±2.43 | 0.00±0.00 | 95.50±4.43 |
| **REM** | 0.03±0.11 | 0.09±0.11 | 0.16±0.28 | 0.00±0.00 | 5.60±2.39 | 91.98±14.50 |
| **Precision** | 99.72±0.76 | 54.68±16.56 | 92.44±8.16 | 70.28±26.70 | 76.23±23.89 | |
| **F1** | 97.18±2.83 | 65.26±13.48 | 87.68±7.55 | 77.88±22.10 | 81.42±21.55 | |
| **Accuracy** = 92.65±4.38 | | **Kappa** = 0.85±0.09 | | **MF1** = 81.88±9.18 | | |

*Table S. 6 Confusion matrix and the result of evaluating proposed method using LOO validation approach for DREAMS dataset using single EEG channel signals.*

|  | Multi-epoch network ($O^*_n$) | | | | | |
|---|---|---|---|---|---|---|
|  | **Wake** | **S1** | **S2** | **SWS** | **REM** | **Recall** |
| **Wake** | 14.42±9.48 | 0.49±0.51 | 0.05±0.08 | 0.03±0.06 | 0.11±0.31 | 93.43±9.48 |
| **S1** | 0.17±0.22 | 6.83±2.67 | 0.33±0.52 | 0.02±0.07 | 0.19±0.22 | 91.02±8.58 |
| **S2** | 0.47±0.58 | 2.45±1.65 | 32.76±8.22 | 3.90±2.74 | 2.24±1.93 | 77.90±7.84 |
| **SWS** | 0.04±0.08 | 0.01±0.02 | 2.13±2.89 | 17.95±6.30 | 0.01±0.02 | 89.54±10.99 |
| **REM** | 0.11±0.29 | 0.26±0.65 | 0.27±0.43 | 0.00±0.00 | 14.75±3.64 | 95.68±7.27 |
| **Precision** | 94.52±4.28 | 68.00±14.35 | 92.29±6.65 | 81.48±15.72 | 86.35±7.57 | |
| **F1** | 93.66±6.19 | 76.58±9.65 | 84.08±5.11 | 83.73±10.85 | 90.32±4.34 | |
| **Accuracy** = 86.71±2.96 | | **Kappa** = 0.82±0.05 | | **MF1** = 85.67±3.94 | | |

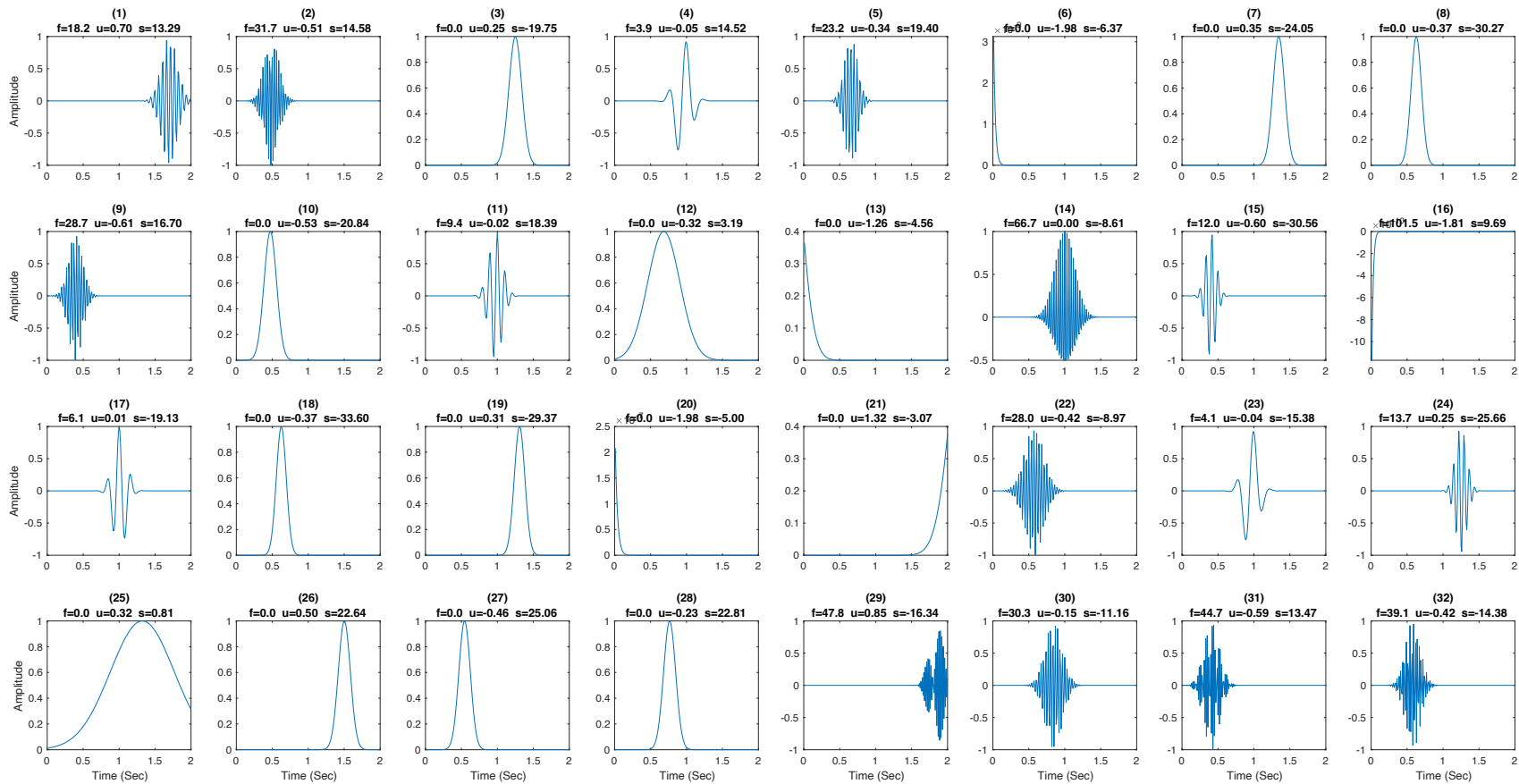

*Figure S. 1. The optimized EEG related Gabor kernels waveforms.*

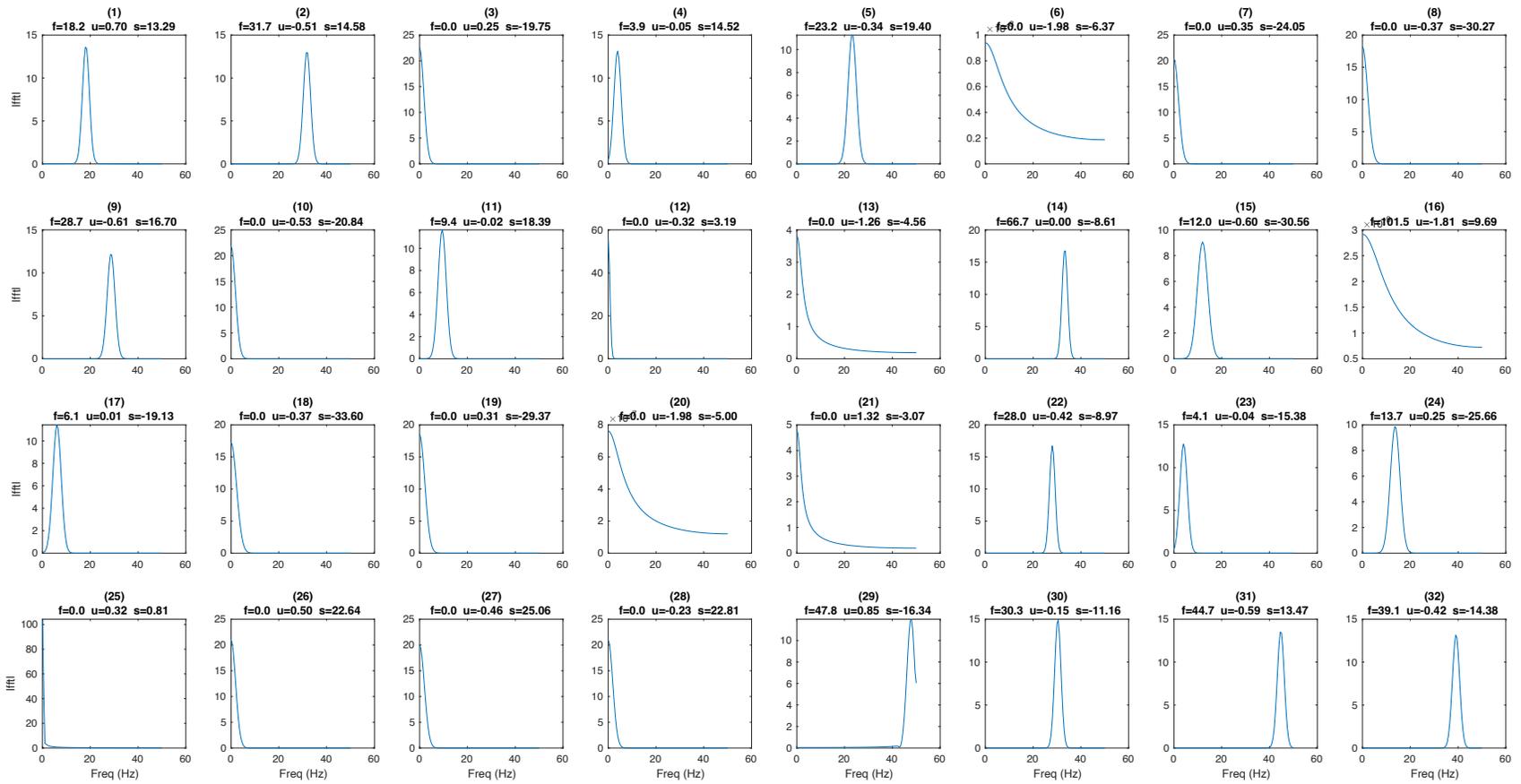

*Figure S. 2. The frequency domain of the optimized EEG related Gabor kernels waveforms.*

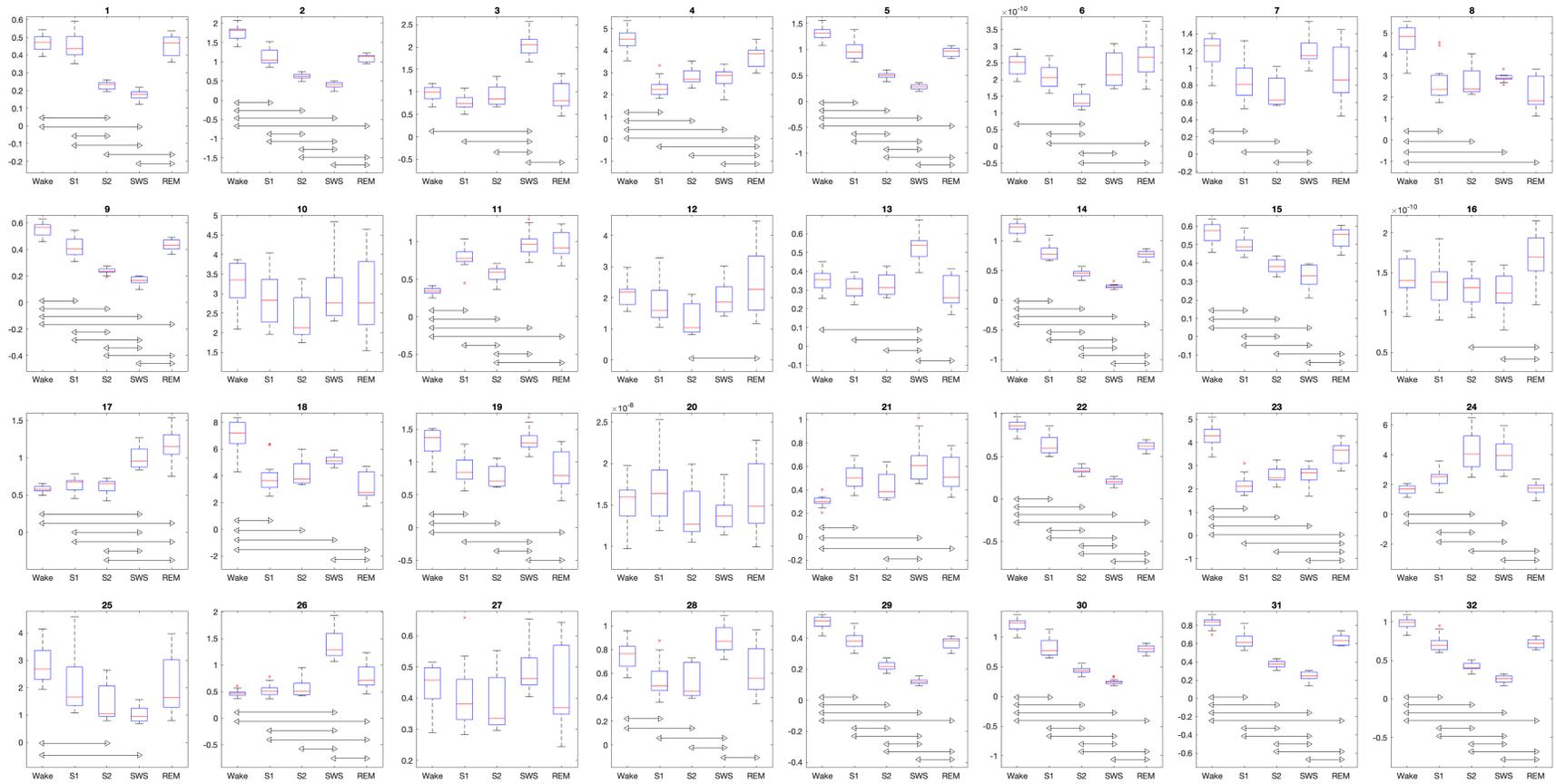

*Figure S. 3. The variation of $\overline{Eff}$ for all Gabor kernels in different stages and the result of the statistical tests. The significant differences (p-value<0.05) are indicated using bidirectional arrows.*

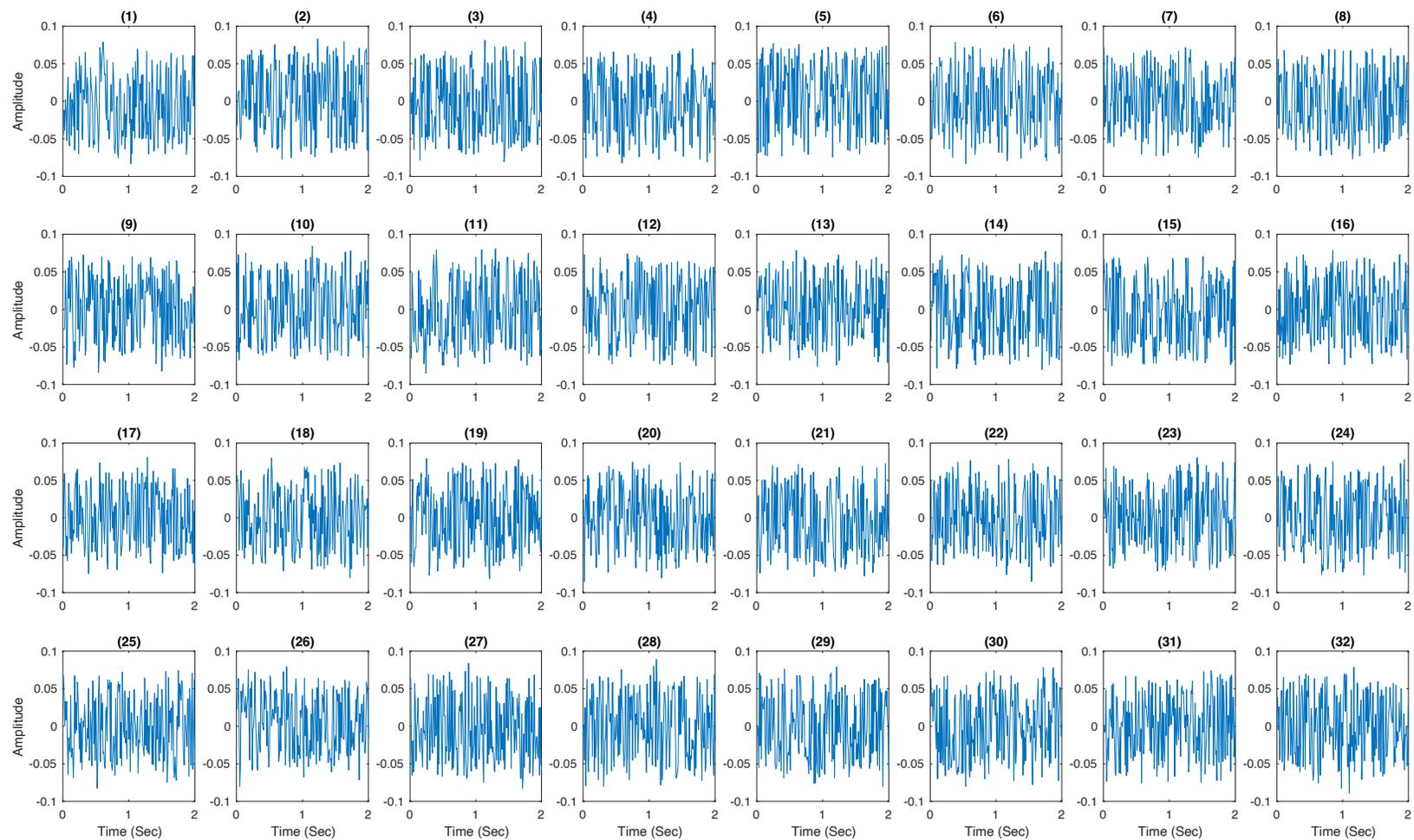

*Figure S. 4 The optimized 1D-convolution filters in the first layer for EEG signals which were replaced after removing Gabor kernels.*